%% file: main.tex
\documentclass[11pt]{article}
\usepackage{fullpage,times}
\usepackage{parskip}

\usepackage{amsmath, amsthm, amssymb, amsfonts, mathtools, graphicx, enumerate}
\usepackage{adjustbox}
\usepackage[utf8]{inputenc} 
\usepackage[T1]{fontenc}    
\usepackage{url}            
\usepackage{booktabs}
\usepackage{nicefrac}       
\usepackage{microtype}      
\usepackage{xspace}
\usepackage{algorithm,algorithmic}
\usepackage{color}
\usepackage{enumitem}
\usepackage{comment}
\usepackage{bm}
\usepackage{apptools}
\usepackage[page, header]{appendix}
\usepackage{titletoc}
\usepackage{array}
\usepackage{varwidth}

\usepackage[scaled=.9]{helvet}

\usepackage{url}
\usepackage{caption}
\usepackage[table,dvipsnames]{xcolor}
\usepackage[colorlinks=true, linkcolor=blue, citecolor=blue]{hyperref}

\usepackage{natbib}
\bibpunct{(}{)}{;}{a}{,}{,}

\usepackage{lipsum}  
\makeatletter
\newcommand{\printfnsymbol}[1]{%
  \textsuperscript{\@fnsymbol{#1}}%
}
\usepackage{style}

\title{Operator Deep Q-Learning: Zero-Shot Reward Transferring in Reinforcement Learning}
\date{}

\author{
  Ziyang Tang\\
  University of Texas at Austin\\
	\texttt{ztang@cs.utexas.edu} 
  \and 
  Yihao Feng \\
  University of Texas at Austin\\
	\texttt{yihao@cs.utexas.edu} \\
  \and 
  Qiang Liu \\
  University of Texas at Austin\\
	\texttt{lqiang@cs.utexas.edu}
}

\begin{document}

\maketitle

\input{main_only}

\input{tex/appendix}

\end{document}

%% file: main_only.tex
\begin{abstract}
Reinforcement learning (RL) has drawn increasing interests in recent years due to its tremendous success in various applications.
However, standard RL algorithms can only be applied for single reward function, and cannot adapt to an unseen reward function quickly.
In this paper, we advocate a general \emph{operator} view of reinforcement learning, which enables us to directly approximate the operator that maps from reward function to value function.
The benefit of learning the operator is that we can incorporate any new reward function as input and attain its corresponding value function in a \emph{zero-shot} manner.
To approximate this special type of operator, 
we design a number of novel operator neural network architectures based on its theoretical properties.
Our design of operator networks outperform the existing methods and the standard design of general purpose operator network,
and we demonstrate the benefit of our operator deep Q-learning framework in several tasks including reward transferring for offline policy evaluation (OPE) and reward transferring for offline policy optimization in a range of tasks.

\end{abstract}

\input{tex/intro}
\input{tex/background}

\input{tex/method}
\input{tex/design}
\input{tex/discussion}

\input{tex/related}
\input{tex/exp}

\section{Conclusion, Limitations and Social Impacts}
\label{sec:conclustion}

We propose an operator view of reinforcement learning, which enables us to transfer new unseen reward in a zero-shot manner. 
Our operator network directly leverages reward function value into the design,
which is more straightforward and generalized compared with previous methods.
One limitation of our operator q-learning is that we need to get access to a predefined reward sampler, which need human knowledge on the specific task.
Therefore, an important future direction is to generalize our method to  reward-free settings \citep[e.g.][]{jin2020reward}.

Our method improves the transferability of RL and hence advance its application in daily life. A potential negative social impact is our algorithm may fail to consider fairness and uncertainty, if fed  with biased or limited data. An important future direction is to investigate the potential issues here. 
\clearpage
\bibliography{ref}

%% file: tex/intro.tex
\section{Introduction}
Reinforcement learning (RL), especially when equipped with powerful deep neural networks, has achieved remarkable success in domains such as playing games \citep[e.g.,]{mnih2015human,silver2018general, vinyals2019grandmaster}, quadrupedal locomotion \citep{haarnoja2018soft} and autonomous driving \citep{kendall2019learning, bellemare2020autonomous}.
However, the standard RL framework is targeted for expertizing a \textit{single} task.
In real life scenarios we may hope our intelligence agents not only are able to expertize in one single task but can also adapt to  unseen new tasks quickly.

We are interested in  \textit{reward transfer} in RL, 
where learned RL agents need to act optimally under different reward functions, 
while the environment transition dynamics for different tasks remain the same.
Recently, reward transfer in RL has attracted wide attentions.
Existing works \citep[e.g.,][]{schaul2015universal, barreto2017successor, barreto2018transfer, borsa2018universal, barreto2020fast} 
provide frameworks that can leverage the \textit{concept} of reward function into the design of value function.
The concept can be the goal of the reward function, or the linear coefficients of a predefined set of basis functions.
However, all these prior proposed frameworks heavily depend on special assumption on the class of reward function, 
for example, universal value function need to get access to the goal state.
A more general framework that allows transferring on arbitrary rewards is still in demand.

In this paper, we consider directly leveraging the reward function value into the design of the (Q-) value function.
We consider an \textit{operator-view} to map a certain reward function to the value function. 
Instead of learning an approximated value function for a specific reward, we learn the approximated \textit{operator} that maps the reward function to its corresponding value function. 
We term this type of operator as \textit{resolvent operator}, 
following the literature of partial differential equations \citep{yosida1971functional}.
The training of the resolvent operator can be seen as a straightforward extension of Q-learning, 
thus we name our new training algorithm \textit{Opeartor Deep Q-Learning}. 

The main difference between operator Q-learning and standard Q-learning is that we sample different reward functions from a predefined reward sampler during the training phase to fit Bellman equations.
And during the testing phase when an unseen test reward function comes, 
our learned operator can map the test function to its corresponding value function directly which yields the (deterministic) optimal policy in a \textit{zero shot} manner.

To approximate the resolvent operator, we need to seek a universal approximator that can approximate any operator.
Recently, \citet{lu2019deeponet} proposed a general purpose way to represent any nonlinear operator by deep neural networks.
However, the  architecture of the general operator neural networks does not take the special properties of resolvent operators into account. 
To address this problem, we advocate a novel design of the resolvent operator
to satisfy a list of axiomatic theoretical properties of the resolvent operator, hence yielding better practical performance.

Experimental results indicate that our operator deep Q-learning can successfully transfer to an unseen reward in a zero shot manner, and achieve better performance compared to 
existing methods
especially in policy evaluation.

\myparagraph{Main Contribution}
Our main contribution is three-fold: 
Firstly, we propose a unified operator view of reinforcement learning which is connected to various topics in RL such as reward transfer in RL, multi-objective RL and off policy evaluation (OPE);
Secondly, by studying the properties of the resolvent operator, we design novel architectures which outperform the vanilla designs;
Thirdly, we conduct a range of experiments to strengthen the benefits of our framework.

%% file: tex/background.tex
\section{Background and Problem Setting}
Consider the reinforcement learning (RL) setting where an agent is executed in an unknown dynamic environment.
At each step $t$, the agent observes a state $s_t$ in state space $\Sset$, takes an action based on the current policy
$a_t\sim \pi(\cdot|s_t)$ in action space $\Aset$, 
 receives a reward $r(s_t, a_t)$ according to a reward function $r\colon \Sset \times \Aset \to \RR$,  and transits to the next state $s_t' = s_{t+1}$ according to an \emph{unknown} transition distribution $s_t'\sim p(\cdot|s_t,a_t)$.

We focus on the \emph{offline, behavior-agnostic settings} \citep[e.g.,][]{nachum2019dualdice,zhang2020gendice,levine2020offline}, 
where we have no access to the real environment and 
can only perform  estimations on an 
offline dataset $\D = \{s_i, a_i, s_i', r_i\}_{i=1}^n$ 
collected from previous experiences following the same model dynamics and reward function 
but under different and \emph{unknown} policies. 
In offline RL, we are interested in either \emph{policy evaluation} or \emph{policy optimization}. 
In policy evaluation, we are interested in estimating the 
the (Q-)value function of a given policy $\pi$ of interest,  
\begin{equation}\label{eqn:defineq}
    q_{\pi,r}(s,a) = \E_{\tau \sim \pi}
    \left [\sum_{t=0}^{\infty} \gamma^t r(s_t, a_t)|s_0 = s, a_0 = a \right ]\,,
\end{equation}
where $\tau = \{s_t, a_t\}_{t=0}^{\infty}$ is the trajectory following policy $\pi$ and $\gamma \in (0,1)$ is a discount factor. 
In policy optimization, we want to get the maximum value function  $q_{*,r}$ among all the possible policies 
\begin{equation*}
    q_{*,r}(s,a) = \max_\pi q_{\pi, r}(s,a)\,.
\end{equation*} 
Both $q_{\pi, r}(s,a)$ and $q_{*, r}(s,a)$ are uniquely characterized by Bellman equation: 
\begin{align}
q_{\pi,r}(s,a) =& r(s,a) + \gamma \Ppi[q_{\pi,r}](s,a) \label{eqn:bellman_qpi}\,,\\
q_{*,r}(s,a) =& r(s,a) + \gamma \Pmax[q_{*,r}](s,a) \label{eqn:bellman_qmax}\,,
\end{align}
where $\Ppi$ and $\Pmax$ are the operators defined as:
\begin{align}
    \Ppi[f](s,a) =& \E_{(s',a')\sim p_\pi(\cdot|s,a)} [f(s',a')]\,, \label{equ:ppidef}\\
    \Pmax[f](s,a) =& \E_{s'\sim p(\cdot|s,a)} [\max_{a'\in \Aset} f(s',a')]\,, \label{equ:pmaxdef}
\end{align}
where $p_\pi(s',a'|s,a) = p(s'|s,a)\pi(a'|s')$. 
By approximating $q_{\pi,r}$ and $q_{*,r}$ with parametric functions (such as neural networks) and 
empirically solving the Bellman equation, we 
obtain the fitted Q Evaluation (FQE) for policy evaluation
and fitted Q Iteration (FQI) (or Q-learning) for policy optimization.

%% file: tex/method.tex
\section{Reward Transfer with Operator Q-Learning}

\begin{algorithm*}[t] 
\caption{(Offline) Operator Q-Learning}  
\label{alg:operator_q_learning_main}
\begin{algorithmic} 
\STATE {\bf Input}: 
Reward distribution $\rdist$;
Offline Dataset $\D$;
Step size $\varepsilon$, $\alpha$.
\STATE {\bf Initial}:
Network parameter $\theta$ and its parameter of target network $\theta' = \theta$.
\REPEAT
\STATE {\bf Update:}
    \STATE 1. Get transition data $\D_n = \{s_i,a_i,s_i'\}_{i=1}^n$ uniformly random from the offline dataset $\D$.
    \STATE 2. Sample a reward function $r\sim \rdist$.
    \STATE 3. Compute the target 
    $
        ~~\mathcal{Y}_{\theta'} [r](s_i, a_i) \gets r(s_i,a_i) + \gamma 
         \max_{a'\in \Aset}\G_{\theta'}[r](s_i',a'),
        ~~\forall i\in[n]\,.
    $
    \STATE 4. Update
    $
        ~\theta \gets \theta - \varepsilon \nabla_{\theta} \L(\theta, \D_n)\,,
    $
    where 
    $
         \L(\theta, \D_n) = \sum_{i=1}^n \left(\mathcal{Y}_{\theta'} [r](s_i,a_i) - \G_\theta [r](s_i,a_i)\right)^2\,.
    $ 
    \STATE 5. Update the target
    $
        ~~\theta' \gets (1-\alpha)\theta' + \alpha \theta.
    $
\UNTIL{convergence} 
\end{algorithmic} 
\end{algorithm*}

In standard RL we assume a fixed reward function $r$, 
which amounts to solving Eq.~\eqref{eqn:bellman_qpi} or Eq.~\eqref{eqn:bellman_qmax}. 
In many practical cases, however, 
the reward functions in the testing environments 
can be different from the rewards collected from the training environments, 
which requires the agents are able to learn reward transferring in RL. 

We approach the reward transferring problem with an \emph{operator-view} on solving the Bellman equations. 
We introduce operator $\G_{\pi}$ and $\G_*$, 
which map an arbitrary reward function $r\colon \Sset \times \Aset \to \RR$ to its corresponding value functions $q_{\pi,r}$ and $q_{*,r}$, that is, 
\begin{align*} 
q_{\pi,r}(s,a) = \G_{\pi}[r](s,a),  && 
q_{*,r}(s,a) = \G_{*}[r](s,a),
\end{align*}
for any $r$ and $(s,a)\in \Sset\times \Aset$. 
We call $\G_{\pi}$ and $\G_*$ the \emph{resolvent operators}, a term drawn from the study of 
partial differential equations. 

We aim to construct approximations $\hat\G_{\pi} \approx \G_\pi$ and $\hat\G_{*} \approx \G_*$ from offline observation, 
by parameterizing them using theoretically-motivated \emph{operator neural network architectures}.  
In this way, when encountering an arbitrary reward function $r_{test}$ during testing, we can directly 
output estimates of the corresponding value functions 
$q_{\pi,r_{test}} \approx \hat\G_{\pi}[r_{test}]$ and $q_{*,r_{test}} \approx \hat\G_{*}[r_{test}]$, 
without addition policy evaluation nor policy optimization for the new testing reward,
hence enabling \emph{zero-shot reward transferring}. 
Essentially, our method aims to solve the whole family of Bellman equations (Eq.~\eqref{eqn:bellman_qpi} and Eq.~\eqref{eqn:bellman_qmax})
for different reward functions $r$, 
rather than a single equation with a fixed $r$ like typical RL settings. 

Our method consists of two critical components: 
1)  theoretically-motivated designs of neural network structures
tailored to approximating the operators $\G_{\pi}$ and $\G_{*}$,
and 2) an algorithm that estimates the parameterized operators from data. 
In the sequel, we first introduce the estimation algorithm, 
which is a relatively straightforward extension of Q-learning.  %
We then discuss the operator neural network design, 
together with theoretical properties of $G_\pi$ and $\G_*$ in Section~\ref{sec:design}.

\myparagraph{Operator Deep Q-Learning}
Plugging $q_{\pi,r} = \G_{\pi}[r]$ and $q_{*,r} = \G_{*}[r]$ into Eq.~\eqref{eqn:bellman_qpi}-\eqref{eqn:bellman_qmax}, we have 
\begin{align*} 
&\G_{\pi}[r](x) = r(x) + \gamma \P_\pi[\G_\pi[r]](x)\,, \\
&\G_{*}[r](x) = r(x) + \gamma \Pmax[\G_*[r]](x),~~~~\forall r, ~x \in \X\,.
\end{align*}

Here for simplicity, we write $x = (s,a)$ and $\X = \Sset \times \Aset$ to denote state, action pair and its corresponding space. 
We learn $\G_\pi$ and $\G_*$ by matching these equations on empirical data. 
Let 
$\mathcal R$ be the distribution 
on a set of reward functions $r\colon \Sset\times \Aset \to \RR$, 
which is expected to cover a wide range of $r$ of potential interest during the training phase.
Let $\G_\theta$ be a parameterized operator with a trainable parameter $\theta$. 
We use $\G_\theta$ to approximate $\G_\pi$ or $\G_*$ by minimizing the expected Bellman loss under $r\sim \mathcal R$: 
$$
\min_{\theta} \E_{x\sim \mathcal D,r\sim \mathcal R}  \left [ [(G_{\theta}[r](x) - r(x) - \gamma 
\hat \P[G_\theta[r]](x))^2 \right ],
$$
where $\hat \P$ denotes the empirical estimation of $\P_\pi$ or $\Pmax$ from the offline dataset. 
For example, $\hat{\P_\pi}[f](x_i) = f(s_i', \pi(s_i'))$.
Similar to fitted Q iteration, 
we propose to iteratively update $\theta$ by 
\begin{equation}\label{eqn:fitted_update}
    \theta_{t+1} \gets 
    \arg\min_{\theta}\{ 
    \E_{x\sim \mathcal D,r\sim \mathcal R} [(\G_\theta[r](x)  - \hat{\mathcal Y}_{\theta_t} [r](x) )^2]\,,\!\!
\end{equation}
where $\hat{\mathcal Y}_{\theta_t} [r](x)  = r(x) + \gamma 
\hat\P[\G_{\theta_t}[r]](x)$ denotes the target. 
See Algorithm~\ref{alg:operator_q_learning_main} for more information. %

Our algorithm differs from the standard DQN in two aspects:
1) We sample (different) random reward functions in each iteration rather than using a fixed reward and
2) we replace the Q-function with $\G_\theta[r]$   %
which generalizes to 
different reward functions $r$. 

%% file: tex/design.tex
    \section{Design of Operator Neural Networks}
\label{sec:design}

\begin{figure}[t]
    \centering
    \includegraphics[width=.85\linewidth]{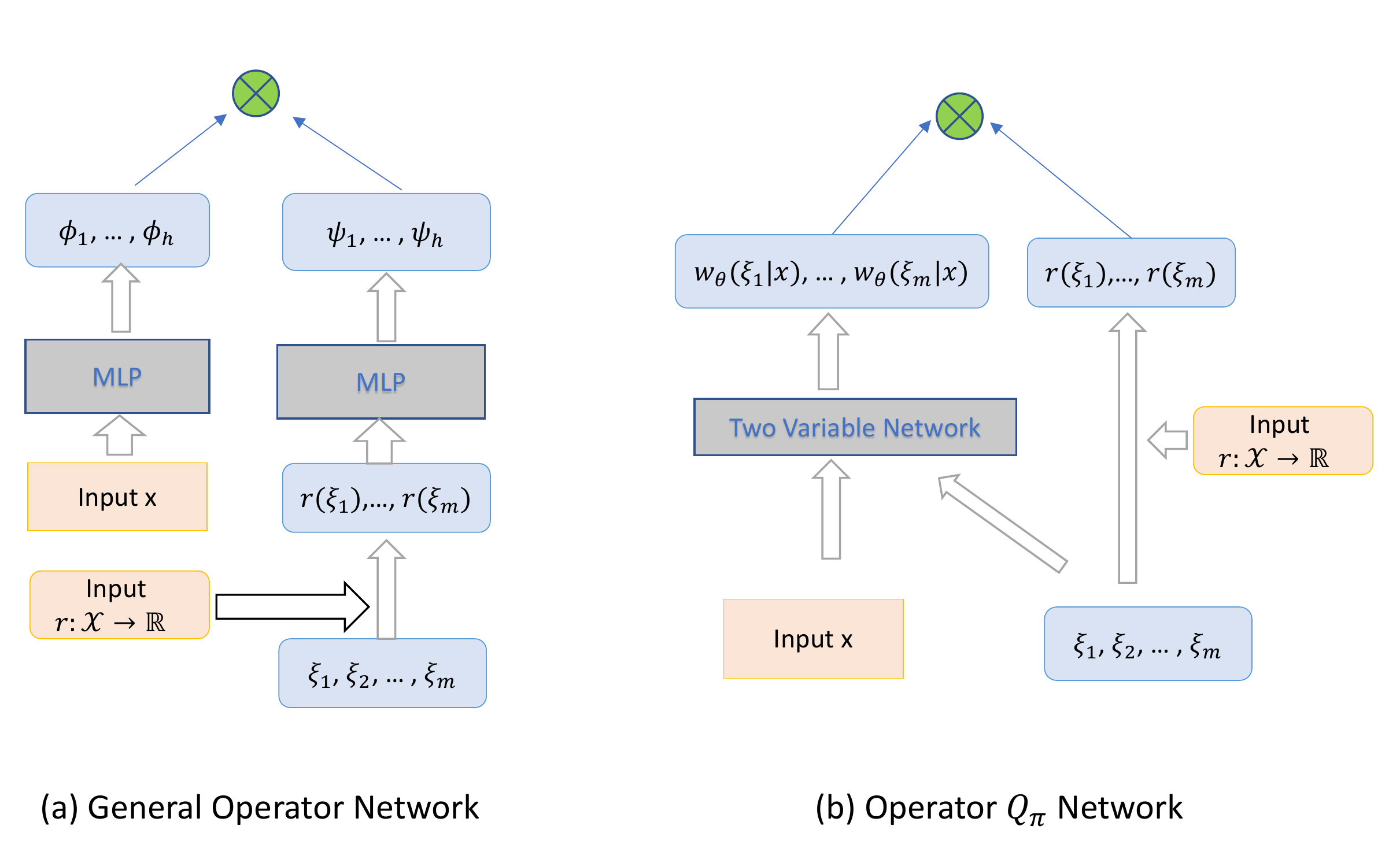}
    \vspace{-1.5em}
    \caption{Illustrations of Network Designs.}
    \label{fig:networks}
\end{figure}

As shown in \citet{lu2019deeponet}, there are general purpose  neural networks (a.k.a \textit{operator neural networks}) that provide universal approximation to general linear and non-linear operators. 
In specific, for any nonlinear operator $\G$ of interest, \citet{lu2019deeponet} approximates $\G[r](x)$ with a \textit{two-stream architecture} of vector value functions $\vv \phi,\vv \psi$ to take input of $r$ and $x$ separately and combine with a dot product:
\begin{equation}\label{eqn:general_purpose_design}
    \widehat{\G}[r](x) = {\vv \phi}(r)^\top {\vv \psi}(x)\,,
\end{equation}
where $\vv \phi$ and $\vv \psi$ are typically parametrized as a general function approximator such as multi-layer perceptron(MLP).
And in particular, since reward function is infinite dimention, 
it is discretized as $r = [r(\xi_1), r(\xi_2),...,r(\xi_m)]^\top$ 
by a set of reference points $\Xi = \{\xi_j\}_{j=1}^m$.
In offline RL settings, we observe rewards $r_i = r(x_i)$ for all $x_i\in \D$,
we can choose (part of) offline data points $\{x_1,\ldots,x_n\}$ as reference points.

However, the general purpose operator network structure in \citet{lu2019deeponet} 
does not leverage the special properties of $\G_\pi$ and $\G_*$ and hence may not generalize well. 
In this section, we propose a theoretically-motivated approach for network design, 
by studying the key theoretical properties of $\G_\pi$ and $\G_*$ 
to guide the design of network structures for policy evaluation and optimization respectively. 

\subsection{Policy Evaluation Case: The Linear Resolvent Operator}
We first consider $\G_\pi$ for policy evaluation, where a fixed target policy $\pi$. It turns out $\G_\pi$ is a linear operator that draws connection to the 
successor representation of \citet{dayan1993improving}. 

\begin{pro}\label{pro:property_gpi}
The resolvent policy evaluation operator
$\Gpi$ is determined by $\P_\pi$ via
\begin{equation}\label{equ:res}
    \Gpi  = \sum_{t=0}^\infty \gamma^t \P_\pi^t = (I-\gamma \Ppi)^{-1},
\end{equation}
which coincides the standard definition of resolvent operator in Markov processes, 
and it satisfies for all reward define on domain $\X$ that: 
    \begin{enumerate}
        \item Linearity: $\Gpi[\alpha r_1 + \beta r_2] = \alpha \Gpi[r_1] + \beta \Gpi[r_2],$ $~\forall \alpha, \beta \in \R, r_1, r_2$. 
        \item Monotonicity: for any non-negative function $\Delta$ such that $\Delta(x)\geq 0,~\forall x\in \X$,
        we have:
        $$
            \Gpi[r+\Delta](x) \geq \Gpi[r](x),~\forall x\in \X.
        $$
        \item Invariant to constant function: $\Gpi[r_C] = r_C/(1-\gamma)$, where $r_C = const$ is a constant function. %
    \end{enumerate}
\end{pro}

From Eq.~\eqref{equ:ppidef}, we have $\P_\pi[r](x)=\E_{x'\sim p_\pi(\cdot|x)}[r(x')],$
which yields 
\begin{align}\label{eqn:point_evaluation}
    \Gpi[r](x) = \frac{1}{1-\gamma}\E_{x'\sim d_\pi(\cdot|x)}[r(x')]\,,
\end{align}
with
\begin{equation}\label{eqn:def_dpi}
    d_\pi(x'|x) =(1-\gamma) \sum_{t=1}^{\infty} \gamma^t p_{\pi}^t(x'|x)\,,
\end{equation}
where $p_\pi^{t}(x
'|x)$ denotes the $t$-step transition probability of the Markov chain when the one-step transition probability is $p_\pi(x'|x)$.  %
 Therefore, $d_\pi(\cdot|x)$ is the discounted average visitation measure of the MDP with policy $\pi$ when initialized from $x$. %
In the tabular case, 
the definition of $d_\pi(\cdot|x)$ coincides with the successor representation in  \citet{dayan1993improving}. 

\myparagraph{Design for Evaluation Operator $\Gpi$}
The representation in Eq.~\eqref{eqn:point_evaluation} sheds important insights for designing operator neural networks for $\Gpi$.  
The expectation over the entire reward function can be approximated by 
a weighted average of the reward value on a finite number of reference points $\Xi = \{\xi_j\}_{j=1}^m$, 
$$
\E_{x'\sim d_\pi(\cdot|x)}[r(x')] 
\approx \sum_{j=1}^m  w_\theta(\xi_j|x) r(\xi_j)\,,
$$
where the reference points $\xi_j$ can be taken (partly) from offline dataset $\xi_j = x_{\sigma(j)}$ 
with random permutation $\sigma\colon \{1,2,\ldots,n\} \to \{1,2,\ldots,n\}$ and when $m = n$, we leverages the whole offline dataset as our reference points.
In this way, we can approximate $\Gpi$ by 
\begin{equation}\label{eqn:design_gpi}
    \G_\theta[r](x) = \frac{1}{1-\gamma}\sum_{j=1}^m 
{w}_\theta({\xi_j}~|~x) r(\xi_j)\,.
\end{equation}

There are two different ways of 
designing  the coefficient function $w_\theta({\xi_j}|x)$.
An attention based design is to parametrize $w$ as importance of $\xi$  w.r.t. $x$
\begin{equation}\label{eqn:attention_design_gpi}
    w_\theta({\xi_j}|x) = \frac{\exp(f_{\theta_f}(\xi_j)^\top g_{\theta_g}(x))}{\sum_{k=1}^m \exp(f_{\theta_f}(\xi_k)^\top g_{\theta_g}(x))}\,,
\end{equation}

where $f_{\theta_f}, g_{\theta_g}\colon \X \to \RR$ are neural networks with parameter $\theta_f,\theta_g$ respectively, and the softmax structure guarantees that all the coefficients are positive and the summation of them is 1, in order to satisfy the properties above.

Another design is to parametrize $w$ as linear decomposition of $\xi$ and $x$:
\begin{equation}\label{eqn:linear_design_gpi}
    w_\theta({\xi_j}|x) = f_{\theta_f}(\xi_j)^\top g_{\theta_g}(x)\,.
\end{equation}
This design does not satisfy all the listed properties such as monotonicity and invariant to constant in Proposition~\ref{pro:property_gpi}, because $w$ here can be negative and the summation of the coefficients is not 1.
However, the linear design can achieve faster computation compared with attention based design, and equipped with random feature, it can approximate the attention design.
Please refer Appendix~\ref{sec:discuss_design} for more details.

Both designs can approximate the true $\Gpi$ arbitrarily well once we have sufficiently number of reference points as the following theorem.
\begin{thm}\label{thm:approximate_gpi}
Suppose $\X$ is compact and $r\in \C(\X)$ is a bounded continual function $|r(x)|\leq 1,~\forall x\in \X$.
Then for any $\varepsilon > 0$ we can find sufficiently large $m$ reference points $\Xi = \{\xi_j\}_{j=1}^m$ such that 
\begin{equation}
    |\G_\theta[r](x) - \Gpi[r](x)| \leq \varepsilon,~\forall x\in \X, ~\forall r\in \F_r.
\end{equation}
In the meanwhile, attention based design of $\G_\theta$ satisfies all the properties in Proposition~\ref{pro:property_gpi}.
\end{thm}

The approximation in Eq.~\eqref{eqn:design_gpi} is closely related to self-normalize importance sampling 
in off policy evaluation; see Appendix~\ref{sec:adjoint} for more discussion.
Figure~\ref{fig:networks} summarizes the difference between our design of the operator network in Eq.~\eqref{eqn:design_gpi} and the general purpose design in Eq.~\eqref{eqn:general_purpose_design}.

\subsection{Policy Optimization Case:  Nonlinear Resolvent Operator}

Let us consider the operator $\Gmax$ for policy optimization. 
Unlike $\Gpi$, $\Gmax$ is a non-linear operator due to the non-linearity of $\Pmax$. 
Thus, we cannot follow the identical design $\Gpi$ for $\Gmax$.
However, the network for $\Gpi$ can serve as the building block for network of 
$\Gmax$, suggested from the following results. %
\begin{pro} \label{pro:property_gmax}
The maximum operator $\Gmax$ can always achieve the optimum value among all policies
\begin{equation}\label{eqn:maxpi}
    \Gmax[r](x) = \max_\pi \Gpi[r](x),~\forall r, x \in \X.
\end{equation}
And it satisfies for all reward define on domain $\X$ the following properties:
\begin{enumerate}
    \item Sub-linearity: $\Gmax [r_1 + r_2] \leq \Gmax[r_1] + \Gmax[r_2]$, $\Gmax[\alpha r] = \alpha \Gmax[r]$, $\forall \alpha\in \R, r_1, r_2$.
    \item Monotonicity: for any non-negative function $\Delta$ such that $\Delta(x)\geq 0,~\forall x\in \X$,
    we have:
    $$
    \Gmax [r+\Delta](x) \geq \Gmax[r](x),~\forall x\in \X.
    $$
    \item Invariance to constant function: $\Gmax[\alpha r + r_C] = \alpha \Gmax[r] + r_C/(1-\gamma)$ for constant function $r_C=const$. 
\end{enumerate}
\end{pro}

Based on Eq.~\eqref{eqn:maxpi}
we propose a max-out structure for $\Gmax$ to satisfy all listed properties.

\paragraph{Max-Out Architecture for $\Gmax$} 
As shown in \eqref{eqn:maxpi},  
$\Gmax$ is the maximum of the $\Gpi$ operators with different policies $\pi$. 
To obtain a computationally tractable structure, 
we discretize the max on a finite set of $K$ linear operators $\G_{\theta_k}[r](x) = \sum_{j=1}^m w_{\theta_k}(\xi_j|x)r(\xi_j)/(1-\gamma)$ and taking the max %
\begin{equation}\label{eqn:design_gmax_maxout}
    \G_{\vv\theta}^{max}[r](x) = \frac{1}{1-\gamma} \max_{k\in [K]}  \sum_{j=1}^m w_{\theta_k}(\xi_j|x) r(\xi_j),
\end{equation}
where $\vv\theta = \{\theta_k\}_{k=1}^K$ is the parameter of the network. 
It is easy to show that 
the max-out design \citep{goodfellow2013maxout} satisfies the properties in the all three properties in Proposition~\ref{pro:property_gmax}.  

The max-out structure can be easily modified from the $\Gpi$ operator network by maintaining $K$ different copies.
Picking the maximum among a number of value functions has been studied in many of the existing network designs of (general) value function \citep[e.g.][]{barreto2017successor, barreto2020fast, berthier2020max}.
Different from Generalized Policy Improvement(GPI), the $K$ copies in our max-out structure does not leverage the pretrained policies/operators in the previous tasks, but serve as a network structure that are jointly optimized together. See Appendix~\ref{sec:discuss_design} for more discussion.

%% file: tex/discussion.tex
\subsection{Connection with Successor Feature}
Successor features (SF) \citep[e.g.,][]{barreto2017successor} consider reward functions that are linear combinations of a set of basis functions, i.e. $r_w = w^\top \phi$.
The successor feature $\psi_\pi$ for basis feature $\phi$ satisfies the Bellman equation:
\begin{equation}\label{eqn:sf_bellman}
    \psi_\pi(x) = \phi(x) + \gamma \Ppi[\psi_\pi](x).\,
\end{equation}
Similar to DQN, successor feature can be approximated by neural network and estimated by Fitted Q Iteration:
$$
\theta_{t+1}\gets \arg\min_{\theta} \{ \E_{x\sim \D, a'\sim \pi(\cdot|a')}[\|\psi_\theta(x) - \phi(x) - \gamma \psi_{\theta'}(s',a')\|^2] \}\,.
$$
By leveraging the corresponding successor feature $\psi_{\theta}\approx \psi_\pi$, the action value function $q_{\pi,r_w}$ can be approximated as $q_{\pi,r_w} \approx q_{\theta,r_w} = w^\top \psi_\theta$.

Given a reward function $r$, its corresponding coefficient $w$ is unknown. However, notice that once we are given the offline dataset $\D = \{x_i, r_i = r(x_i)\}$, the linear coefficient $w$ can be estimated by ordinary least square (OLS) with a closed form:
\begin{equation}\label{eqn:ols_w}
    \hat{w} = \left(\hat{\E}_{\D}[\phi(x)\phi(x)^\top]\right)^{-1} \hat{\E}_{\D}[\phi(x)r(x)]
    := \frac{\Sigma_{\phi}^{-1}}{n} \sum_{i=1}^n \phi(x_i)r(x_i),\,
\end{equation}

where $\Sigma_{\phi} := \hat{\E}_{\D}[\phi(x)\phi(x)^\top]$\footnote{For simplicity we assume $\Sigma_\phi$ is invertible, otherwise we can add an $\ell_2$ regularization term on $w$.}.
Plugging it into $q_{\theta, r_w}$, we have:
\begin{equation}\label{eqn:sf_operator_view}
    \hat{w}^\top \psi_\theta =  \frac{1}{1-\gamma}\sum_{i=1}^n \left(\frac{(1-\gamma)\Sigma_{\phi}^{-1}}{n}\phi(x_i)^\top \psi_\theta(x)\right) r(x_i)
    := \frac{1}{1-\gamma}\sum_{i=1}^n w(x_i|x) r(x_i),\,
\end{equation}
where $w(x_i|x) =: \frac{(1-\gamma)\Sigma_{\phi}^{-1}}{n}\phi(x_i)^\top \psi_\theta(x)$ can be linearly decomposed as vector value function with respect to $x_i$ and $x$.
Compared with linear design of $\Gpi$ in Eq.~\eqref{eqn:linear_design_gpi}, we can see that successor feature can be viewed as a special case with a fixed vector function $f(x_i) = \frac{(1-\gamma)\Sigma_{\phi}^{-1}}{n}\phi(x_i)$.

%% file: tex/related.tex
\section{Related Works}

\myparagraph{Universal Value Function and Successor Features}
The notion of \textit{reward transfer} is not new in reinforcement learning, and has been studied in literature.
Existing methods aim to capture a \textit{concept} that can represent a reward function.
By leveraging the concept into the design of the value function network, the universal value function can generalize across different reward functions.
Different methods leverage different concepts to represent the reward function.
Universal value function approximators (UVFA) \citep{schaul2015universal} considers a special type of reward function that has one specific goal state, i.e. $r_g(s,a) = f(s,a, g)$,
and leverage the information of goal state into the design;
Successor features (SF) \citep{barreto2017successor,barreto2018transfer,borsa2018universal, barreto2020fast} considers reward functions that are linear combinations of a set of basis functions, i.e. $r = w^\top \phi$,
and leverage the coefficient weights $w$ into the design.
Both methods rely on the assumption of the reward function class to guarantee generalization.
And typically they cannot get access to the actual concept directly, 
and need another auxiliary loss function to estimate the concept from the true reward value \citep{kulkarni2016deep}.
Our method is a natural generalization on both methods and can directly plug in the true reward value directly.

\myparagraph{Multi-task/Meta Reinforcement Learning}
Multi-objective RL \citep[e.g.][]{roijers2013survey,van2014multi, li2019multi, yu2020gradient} 
deals with learning control policies to simultaneously optimize over several criteria.
Their main goal is not transferring knowledge to a new unseen task, but rather cope with the conflict in the current tasks.
However, if they consider a new reward function that is a linear combination of the predefined criteria functions \citep{yang2019generalized}, e.g. lies in the optimal Pareto frontiers of value function, then it can be viewed as a special case of SF, which is related to our methods.

Meta reinforcement learning \citep[e.g.,][]{duan2016rl, finn2017model, nichol2018first, xu2018meta, rakelly2019efficient, Zintgraf2020VariBAD} can be seen as a generalized settings of reward transfer, where the difference between the tasks can also be different in the underlying dynamics. 
And usually they still need few-shot interactions with the environment to generalize, differ from our pure offline settings.

\myparagraph{Off Policy Evaluation(OPE)}
Our design of resolvent operator $\Gpi$ is highly related to the recent advances of density-based OPE 
methods \citep[e.g.,][]{liu2018breaking, nachum2019dualdice, tang2019doubly,mousavi2019black, zhang2020gendice,zhang2020gradientdice}, see more discussion in Section~\ref{sec:adjoint}.
However, density-based OPE methods usually focus on a fixed initial distribution 
while our conditional density in Eq.~\eqref{eqn:def_dpi} can be more flexible to handle arbitrary initial state-action pairs.

%% file: tex/exp.tex
\section{Experiments}
\label{sec:experiment}

\newcommand{\figsize}{0.14}
\begin{figure}
    \centering
    \includegraphics[width =.95\linewidth]{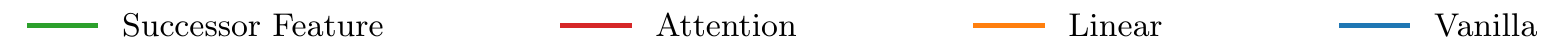}
    \begin{tabular}{cccccc}
    \raisebox{0.5em}{\rotatebox{90}{\tiny Average MSE}}
    \includegraphics[width = \figsize\textwidth]{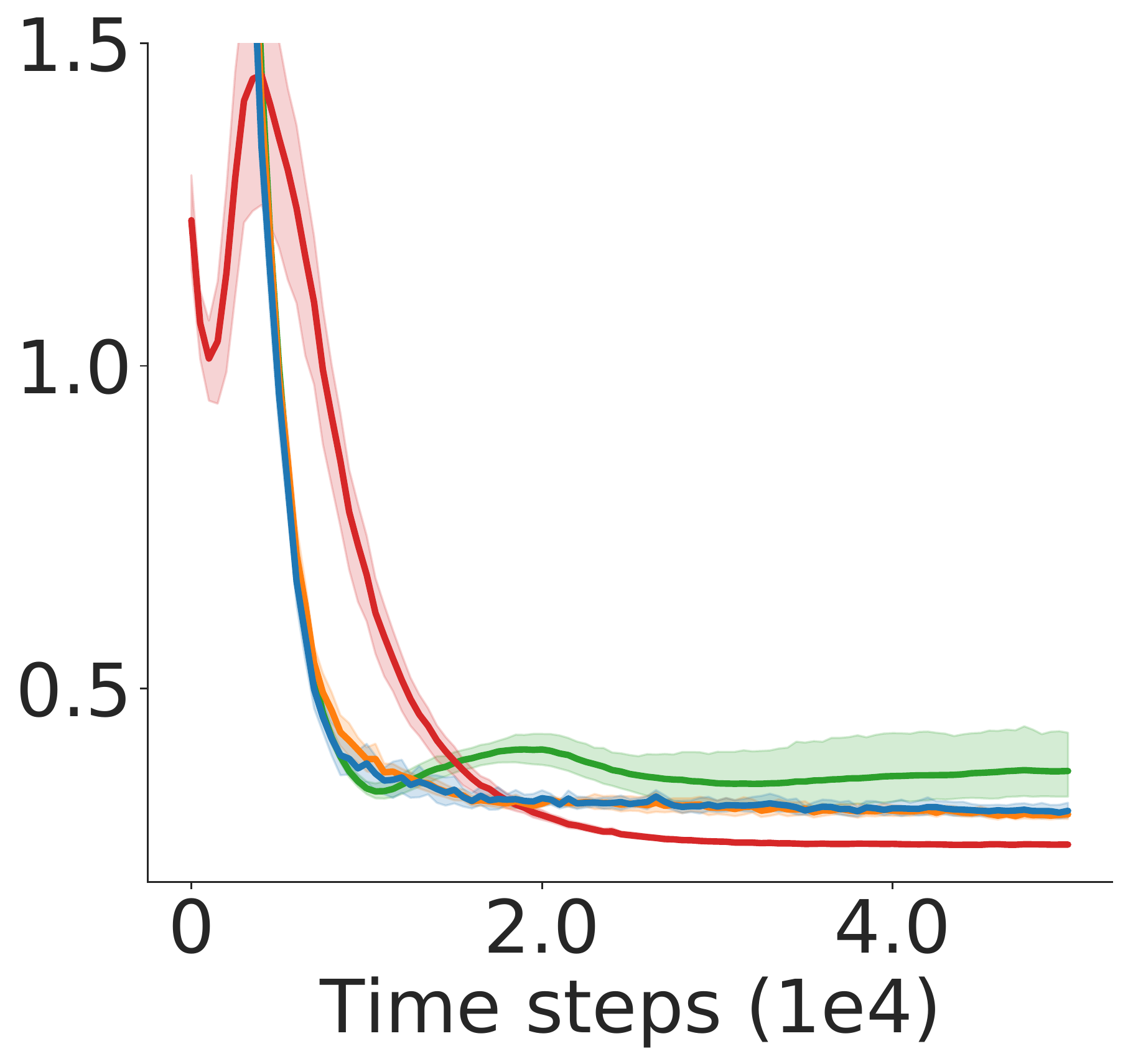} &
    \includegraphics[width = \figsize\textwidth]{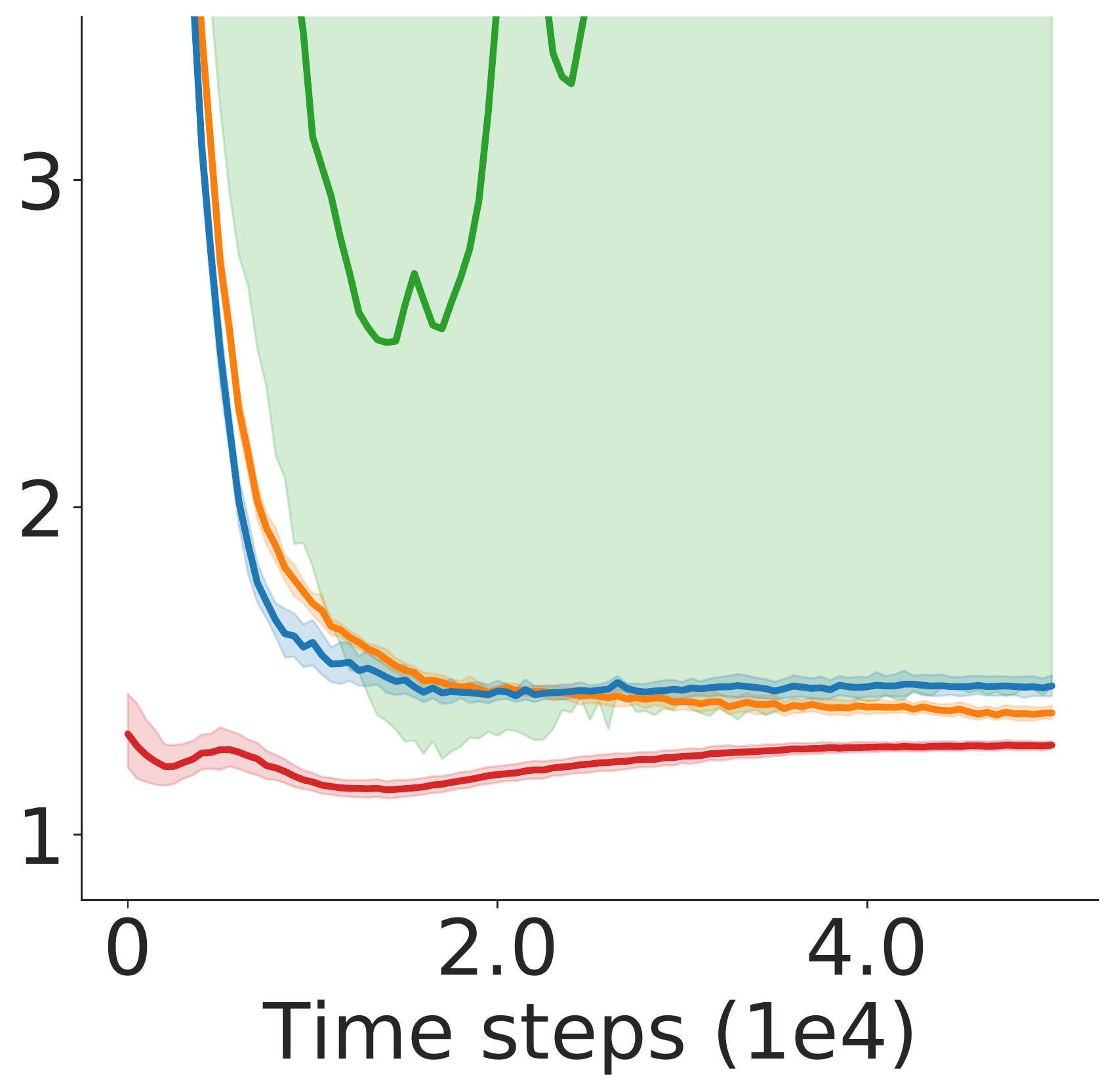} &
    \includegraphics[width = \figsize\textwidth]{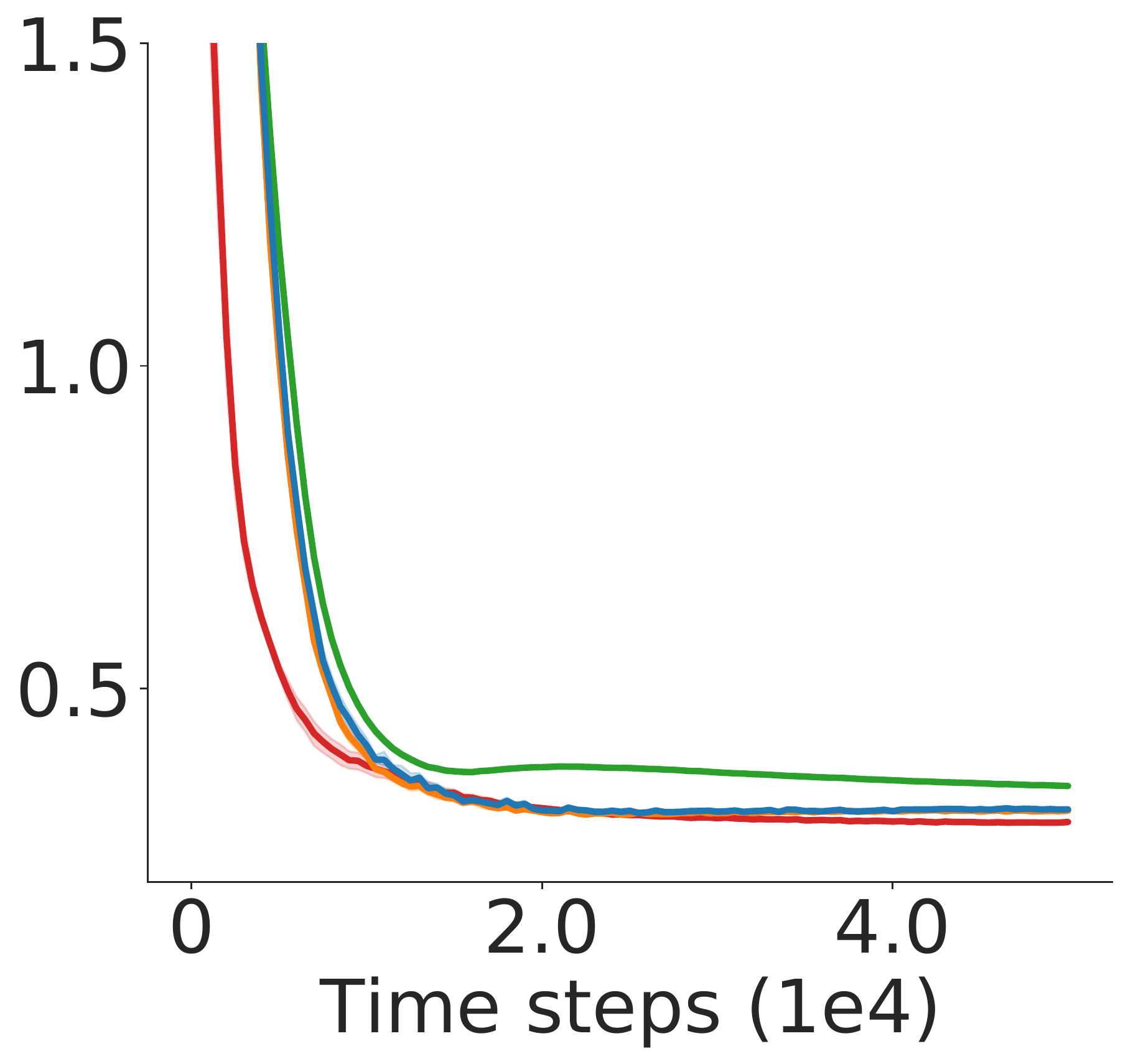} &
    \includegraphics[width = \figsize\textwidth]{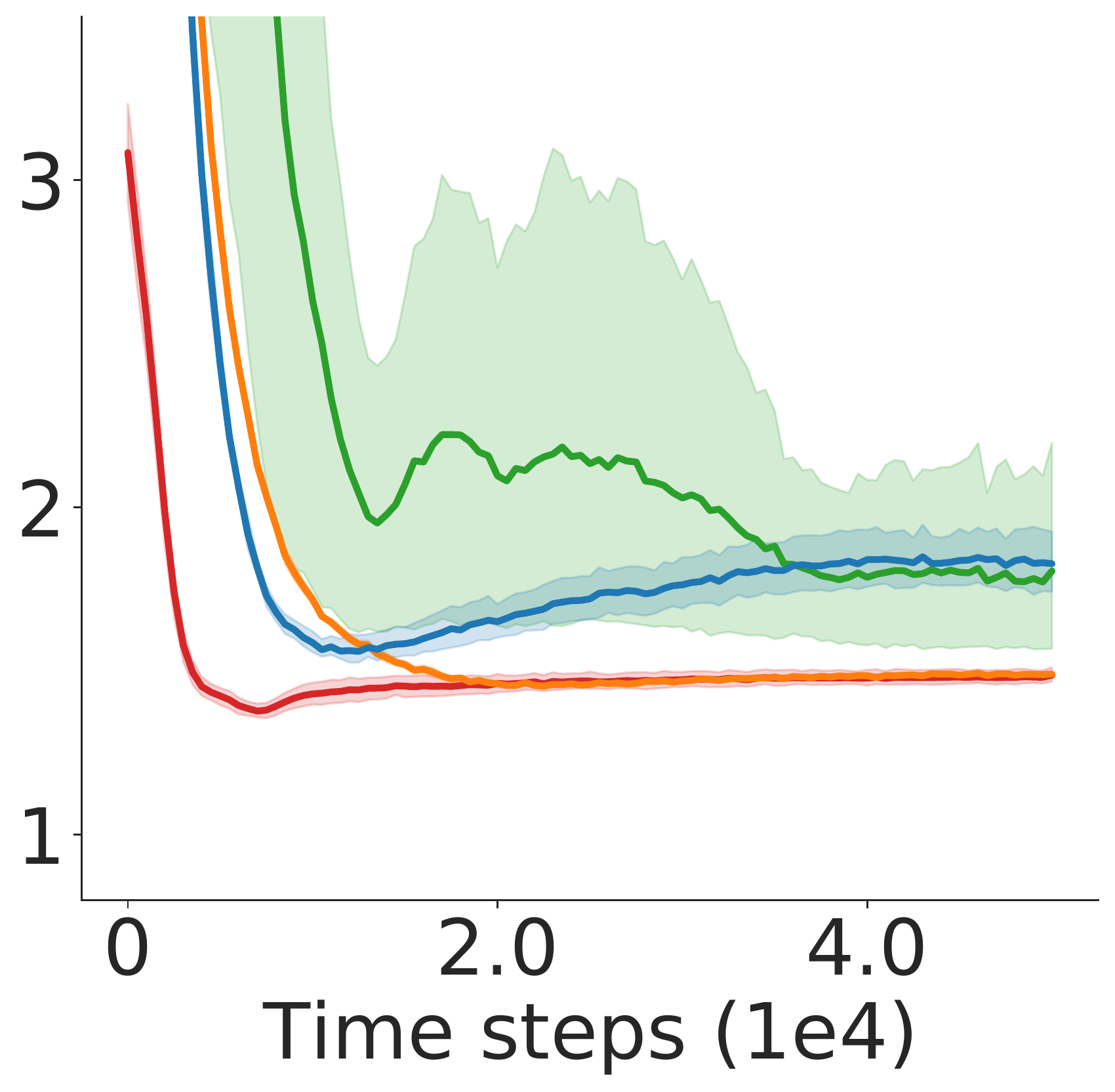} &
    \includegraphics[width = \figsize\textwidth]{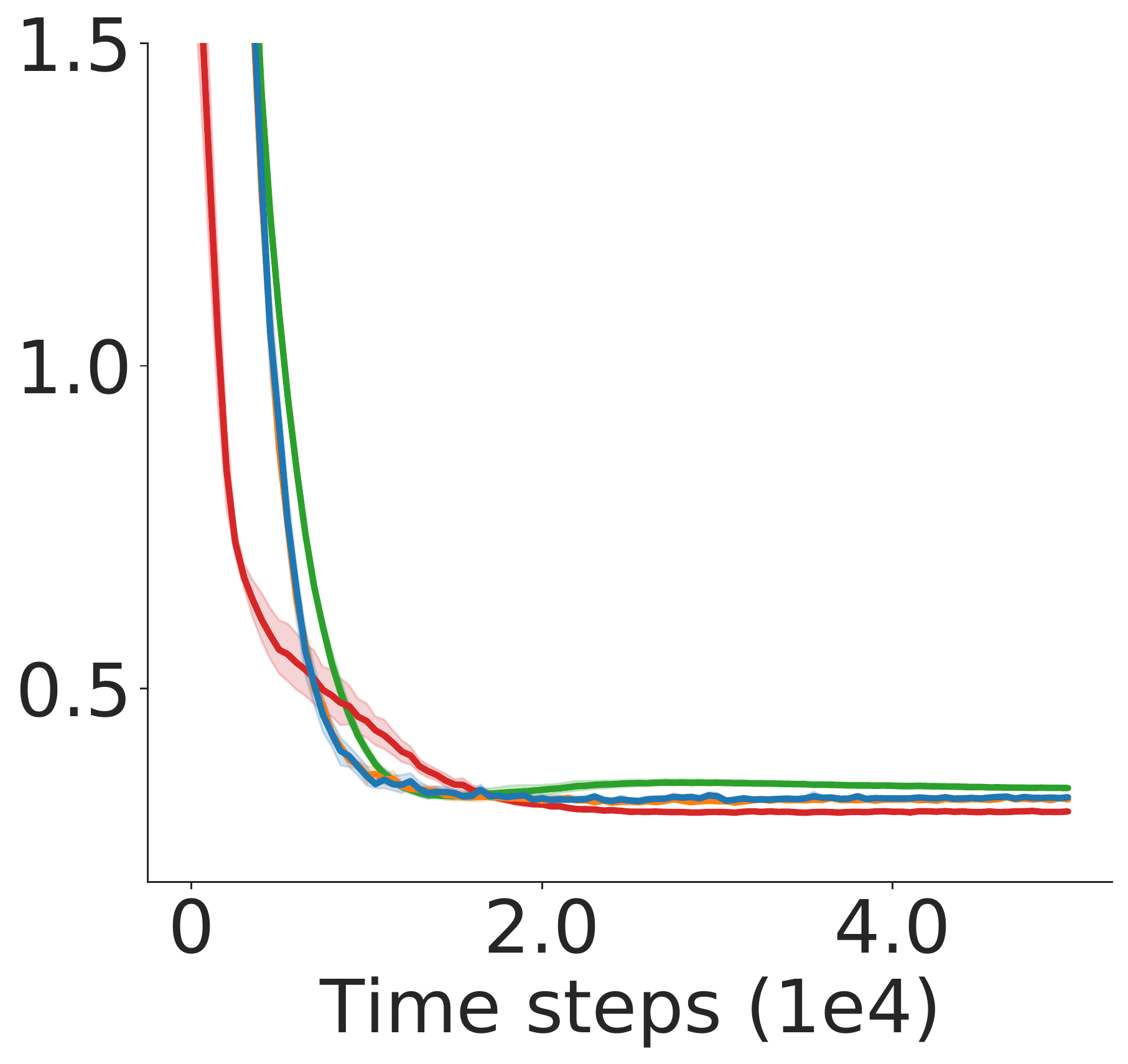} &
    \includegraphics[width = \figsize\textwidth]{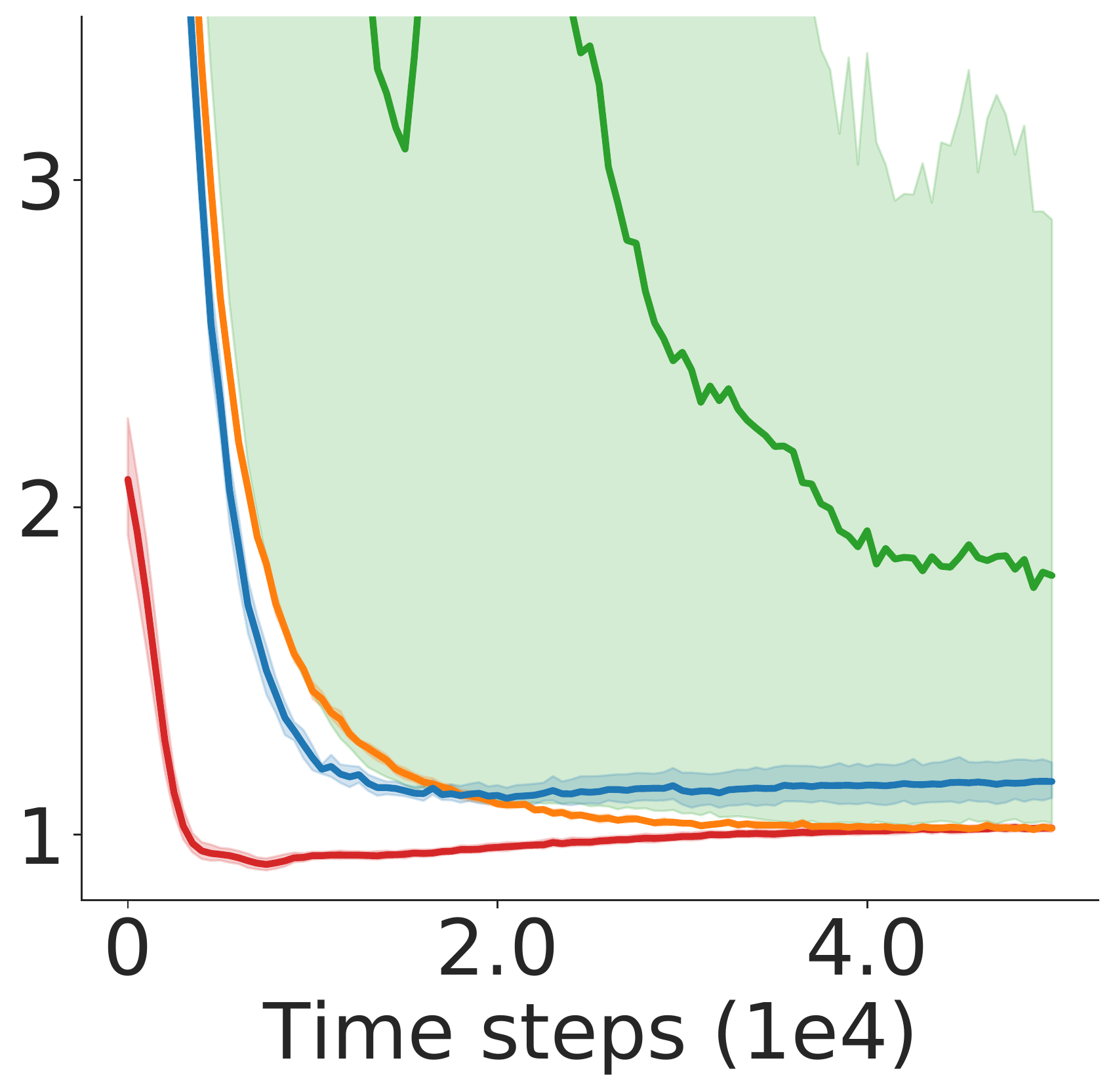} \\
    
    \raisebox{0.5em}{\rotatebox{90}{\tiny Average MSE}}
    \includegraphics[width = \figsize\textwidth]{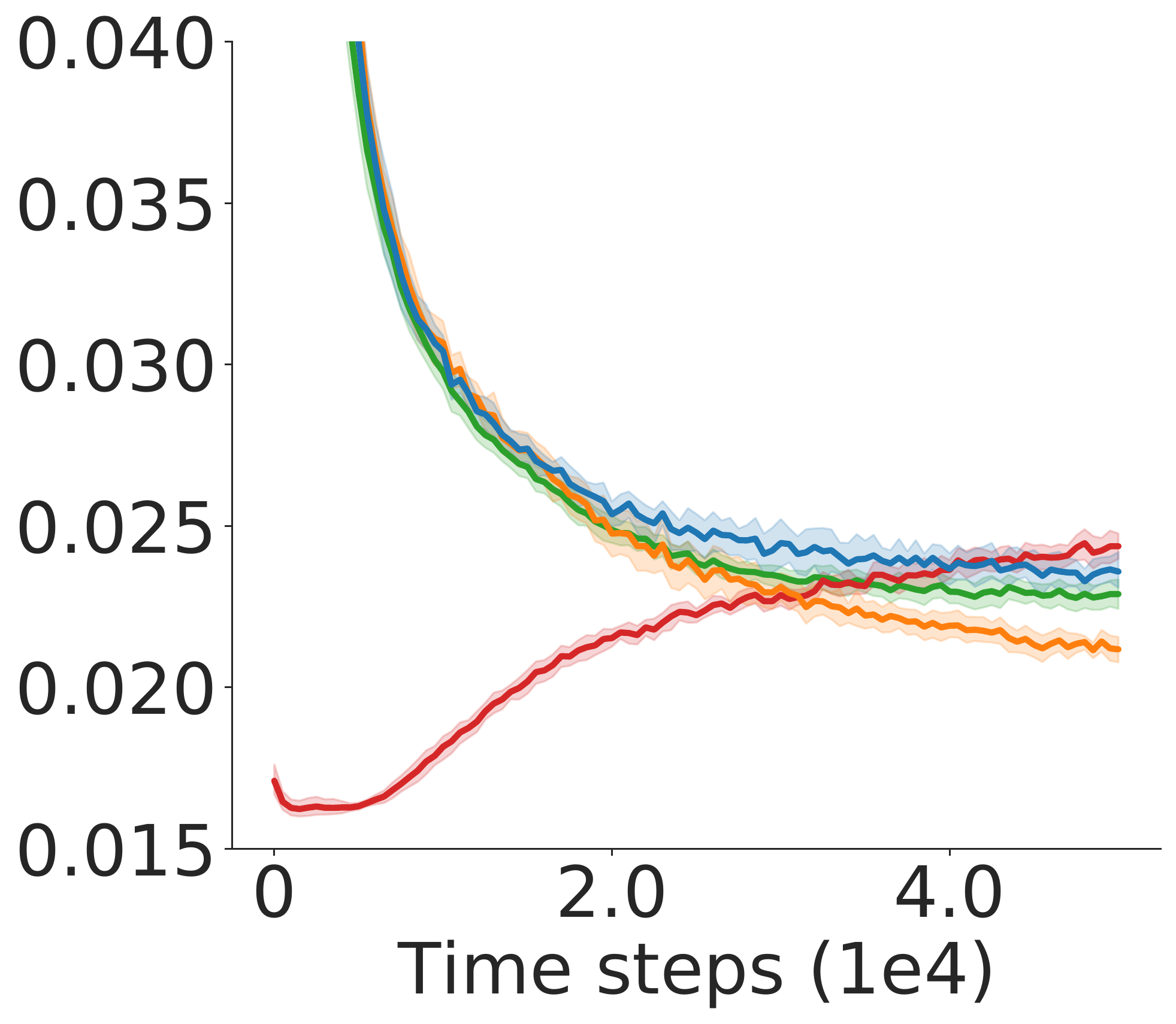} &
    \includegraphics[width = \figsize\textwidth]{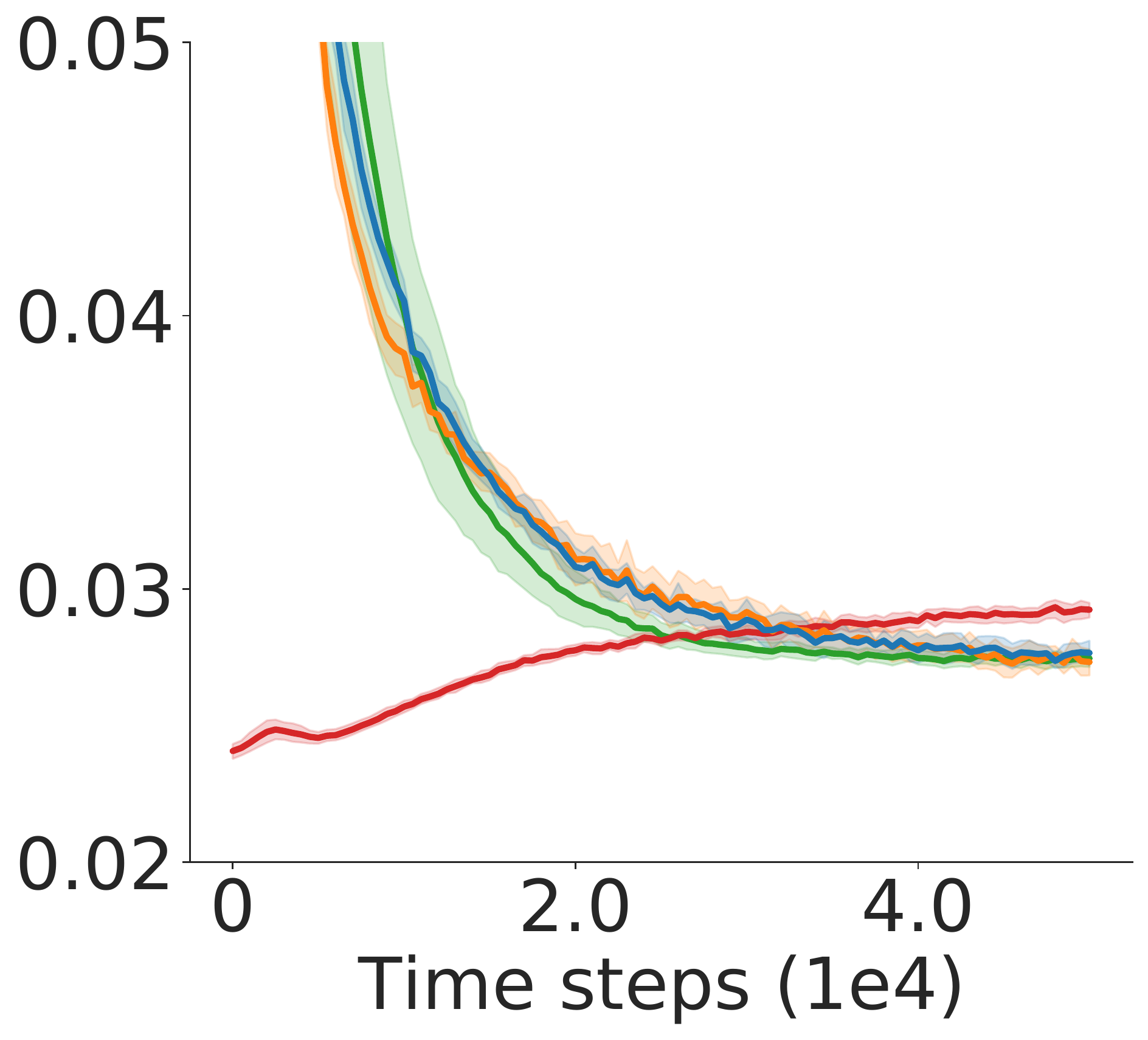} &
    \includegraphics[width = \figsize\textwidth]{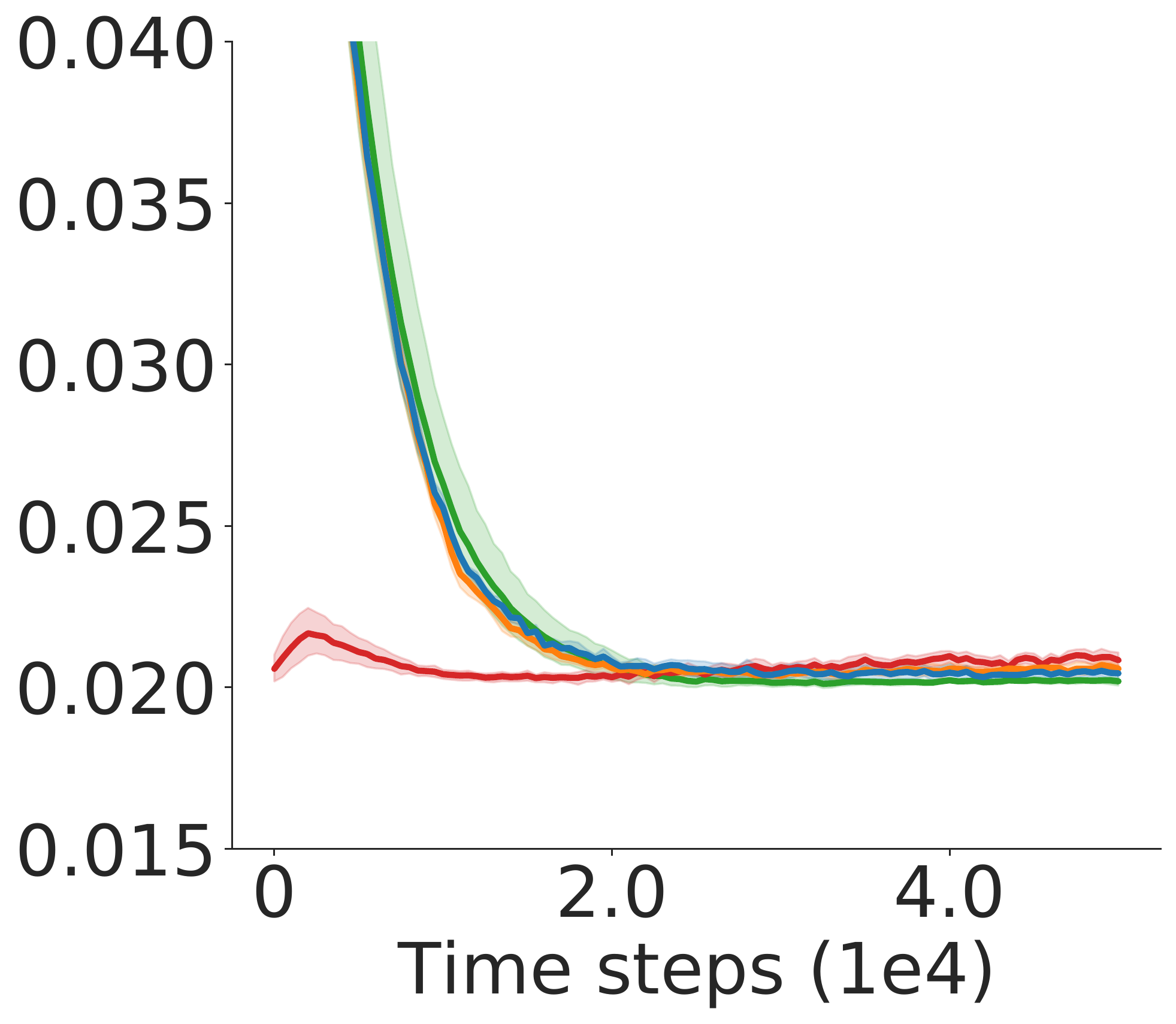} &
    \includegraphics[width = \figsize\textwidth]{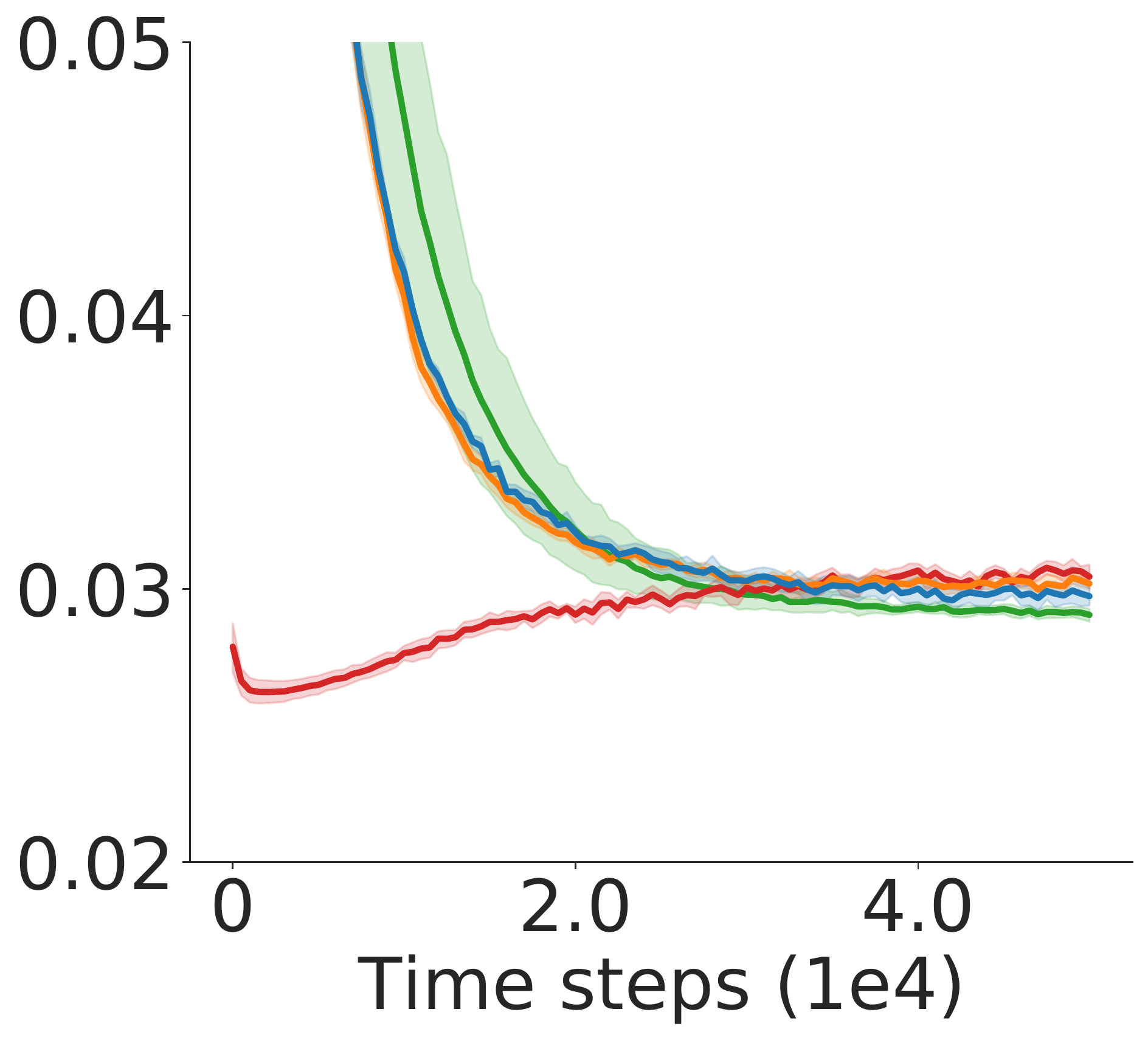} &
    \includegraphics[width = \figsize\textwidth]{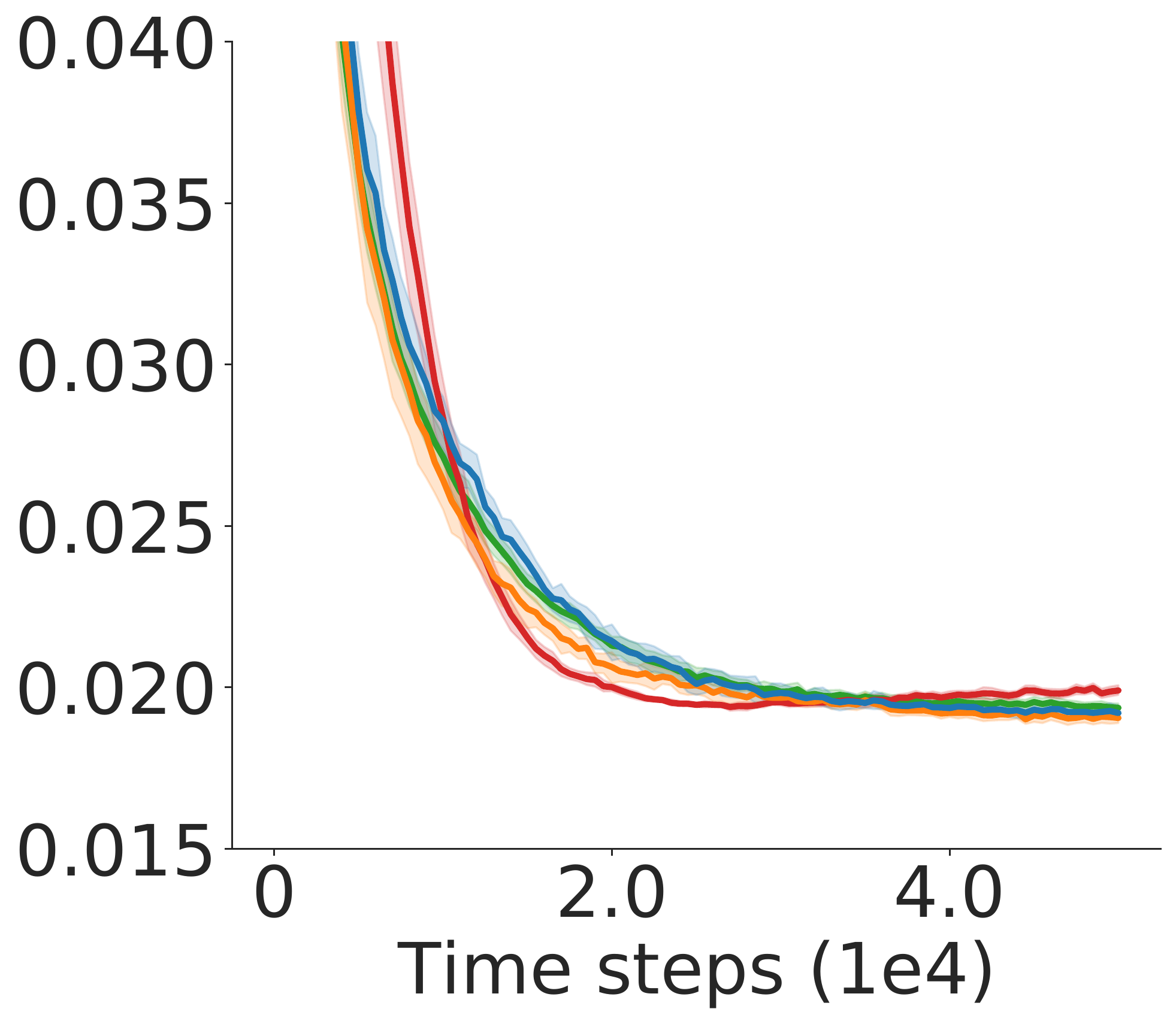} &
    \includegraphics[width = \figsize\textwidth]{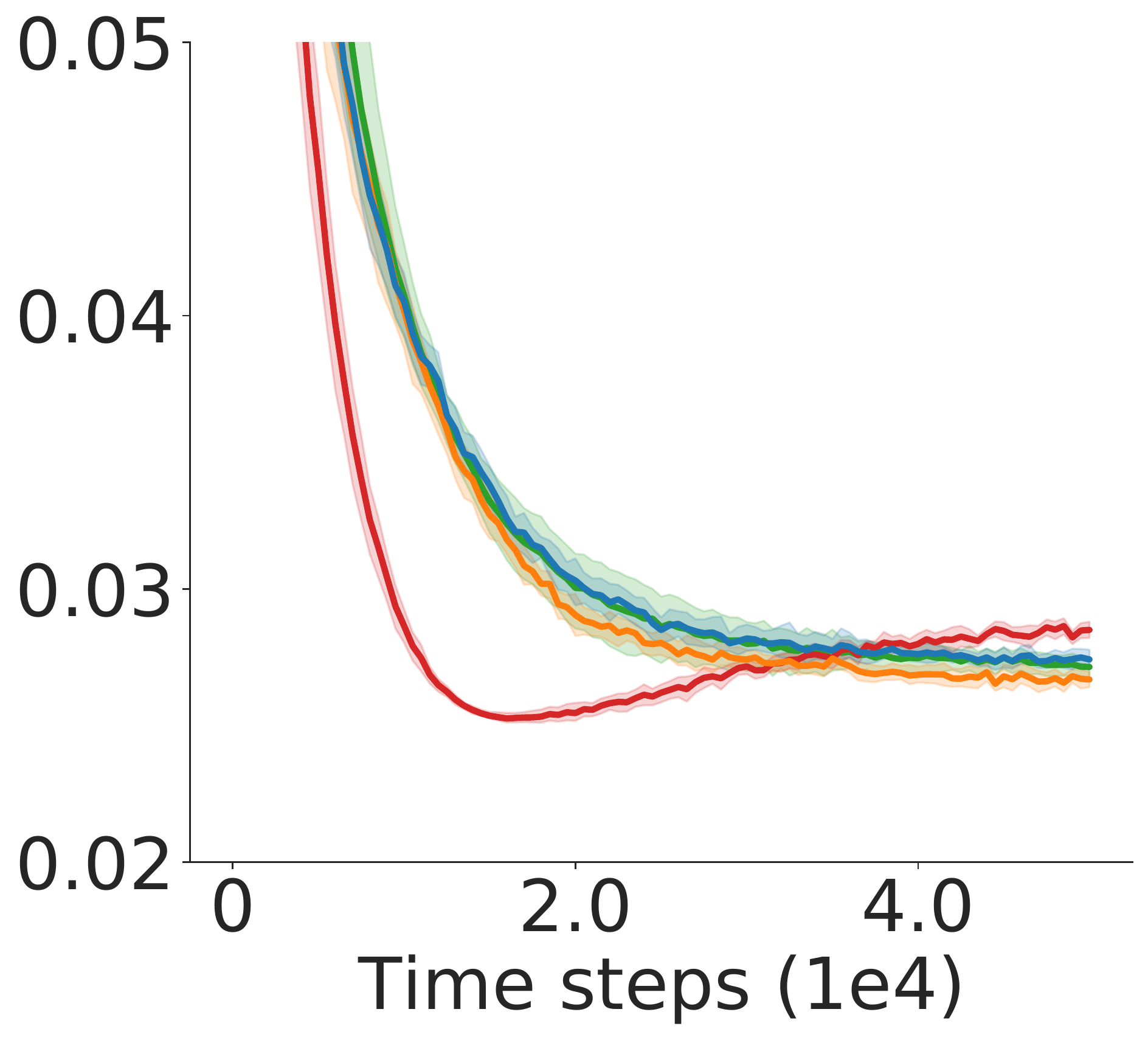} \\
    \raisebox{0.5em}{\rotatebox{90}{\tiny Average MSE}} 
    \includegraphics[width = \figsize\textwidth]{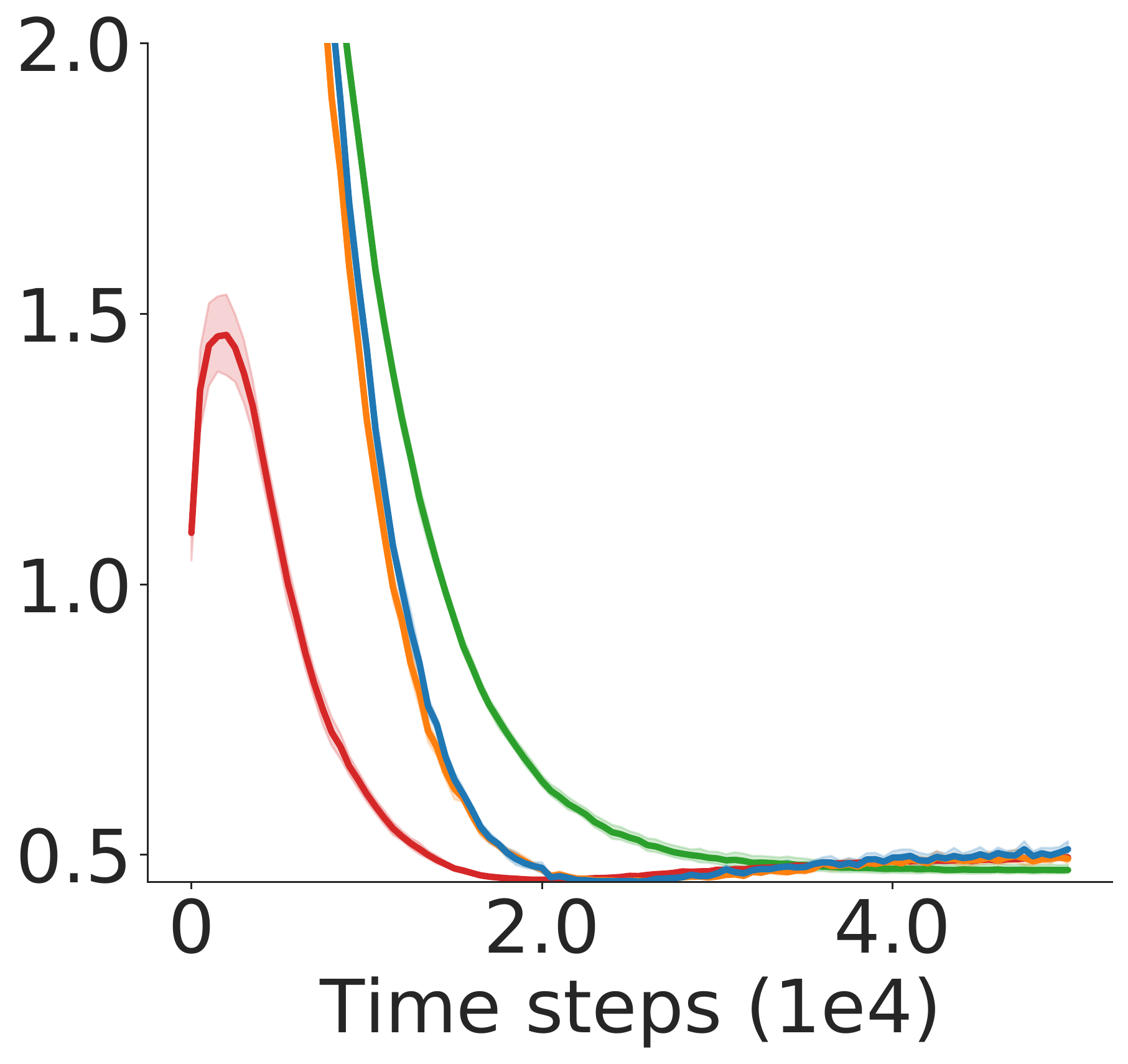} &
    \includegraphics[width = \figsize\textwidth]{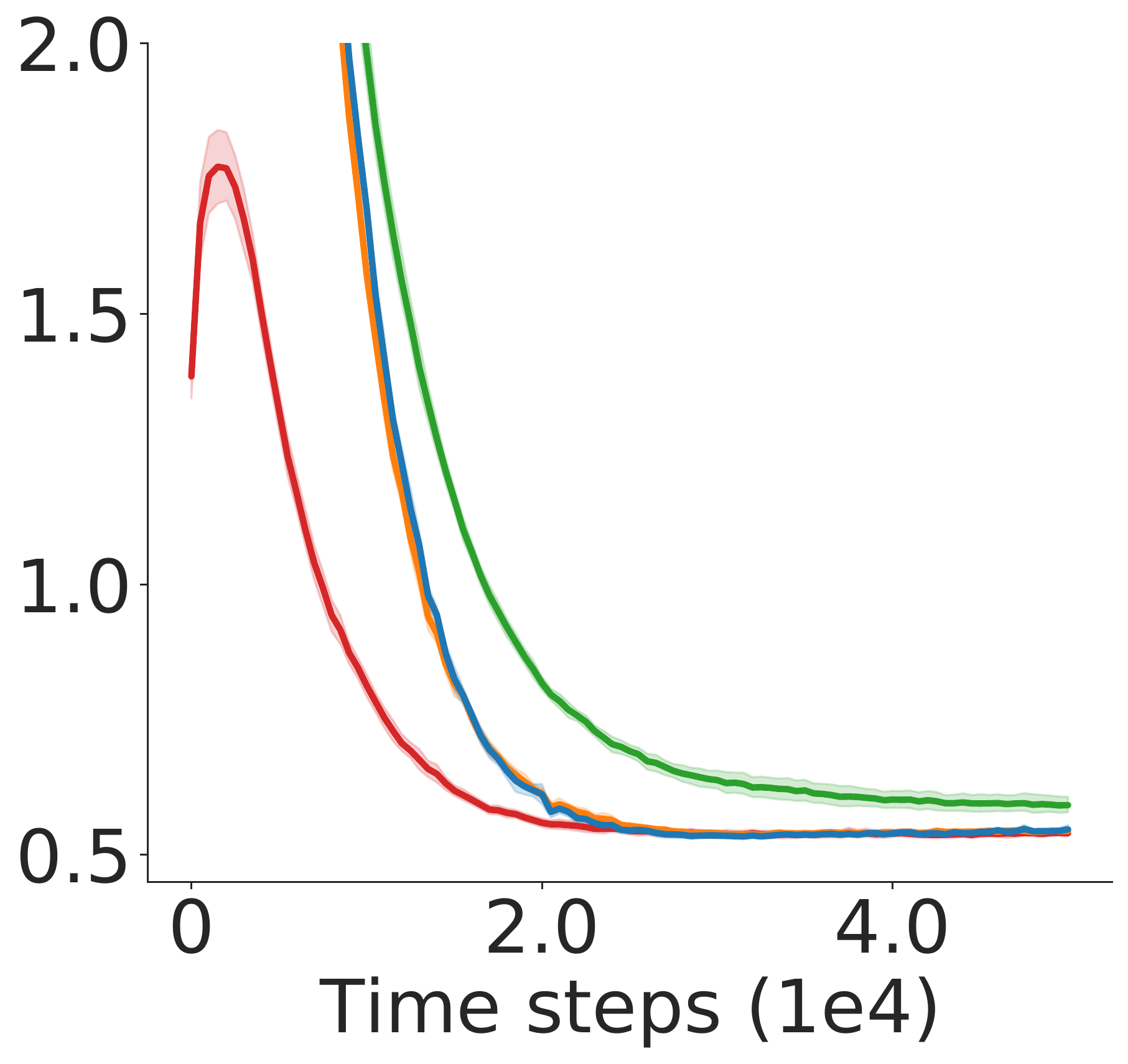} &
    \includegraphics[width = \figsize\textwidth]{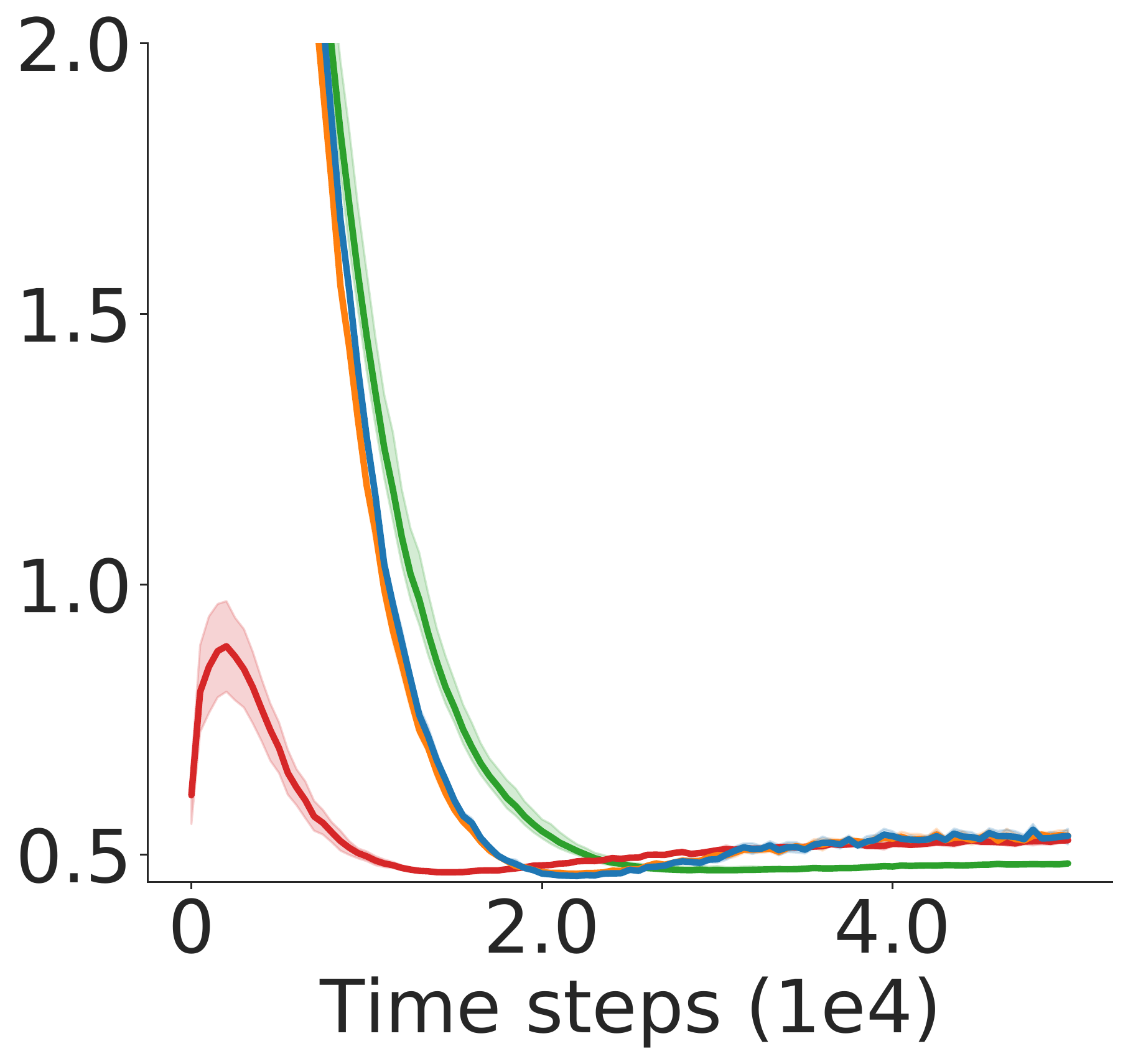} &
    \includegraphics[width = \figsize\textwidth]{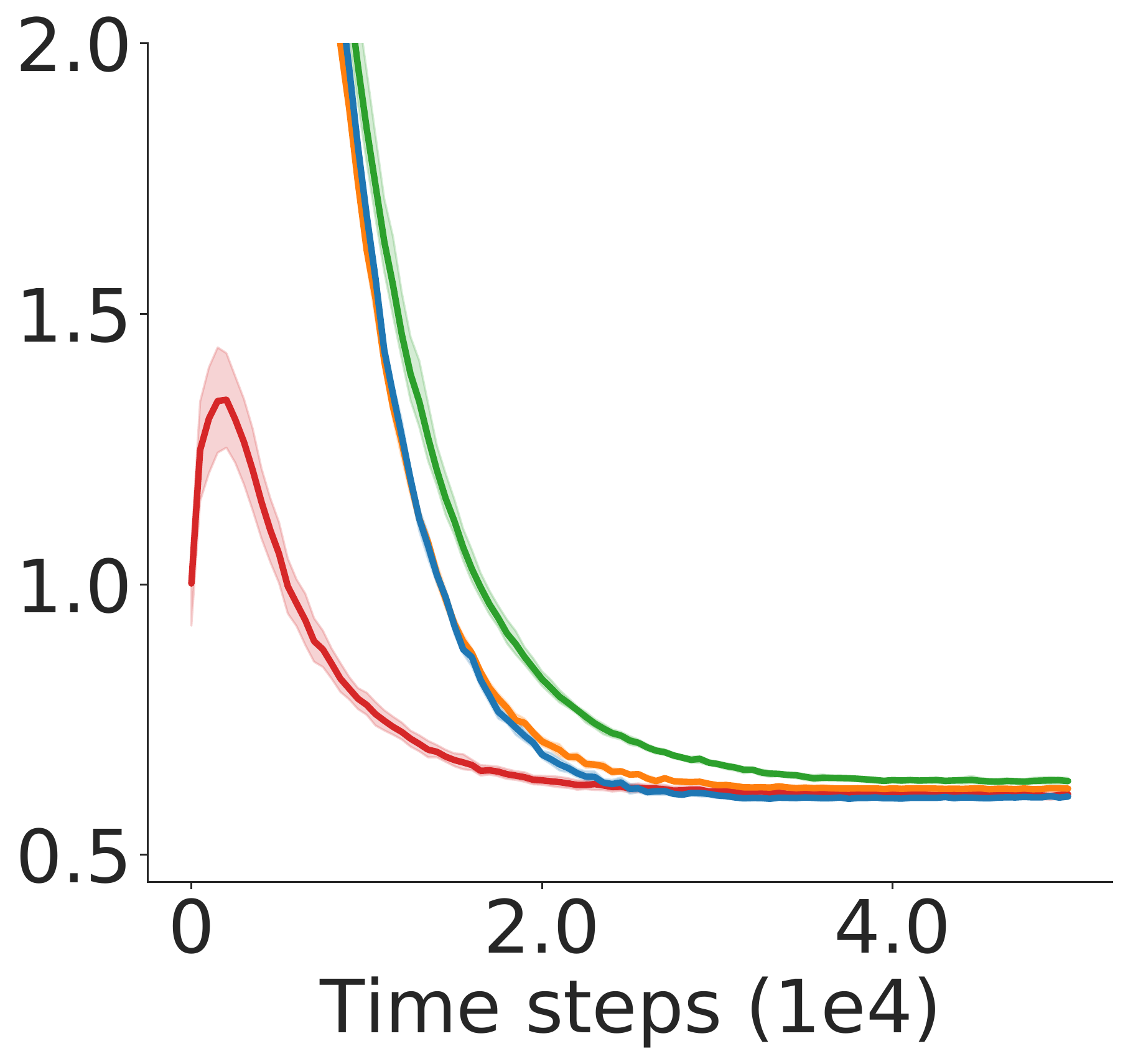} &
    \includegraphics[width = \figsize\textwidth]{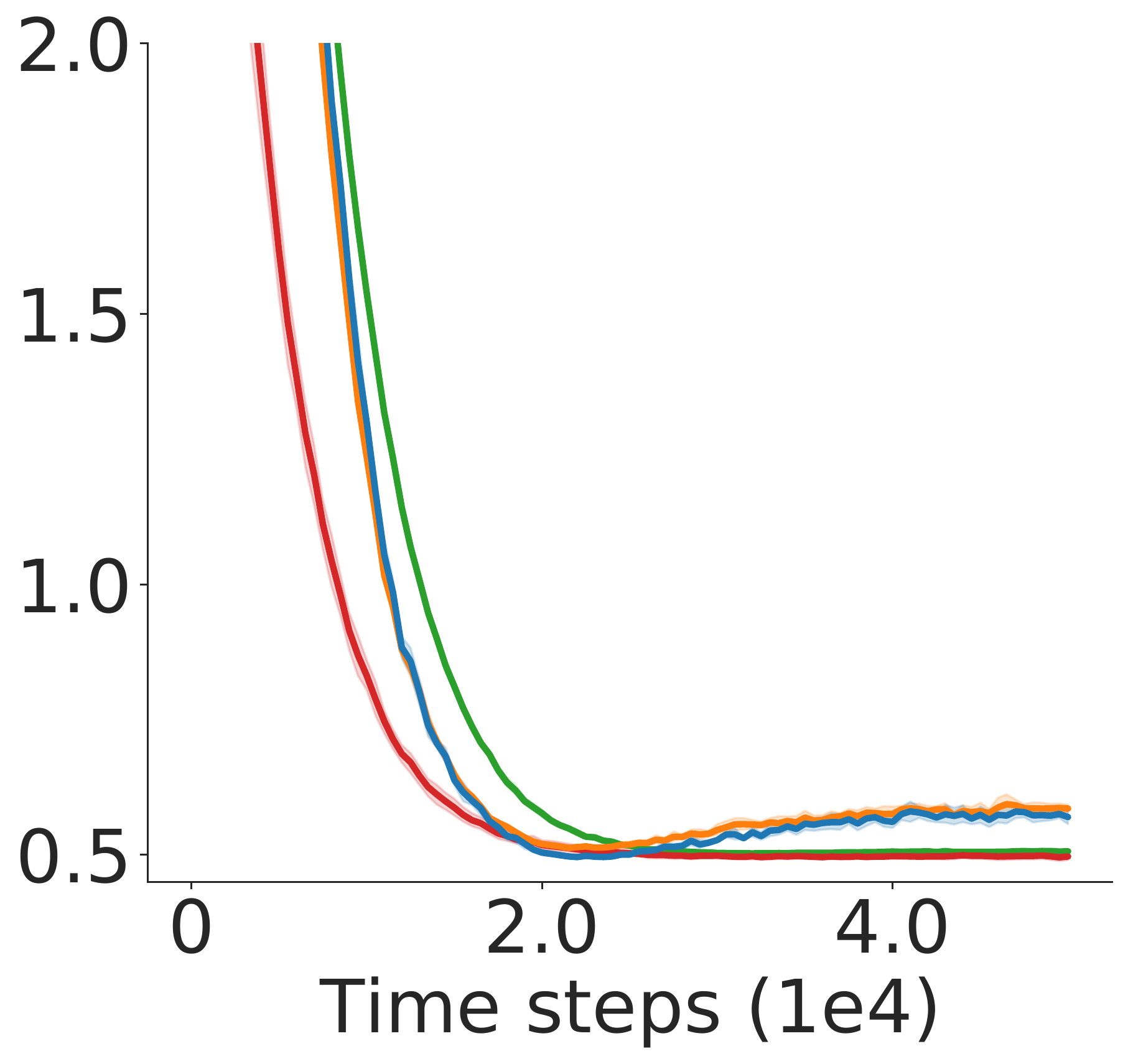} &
    \includegraphics[width = \figsize\textwidth]{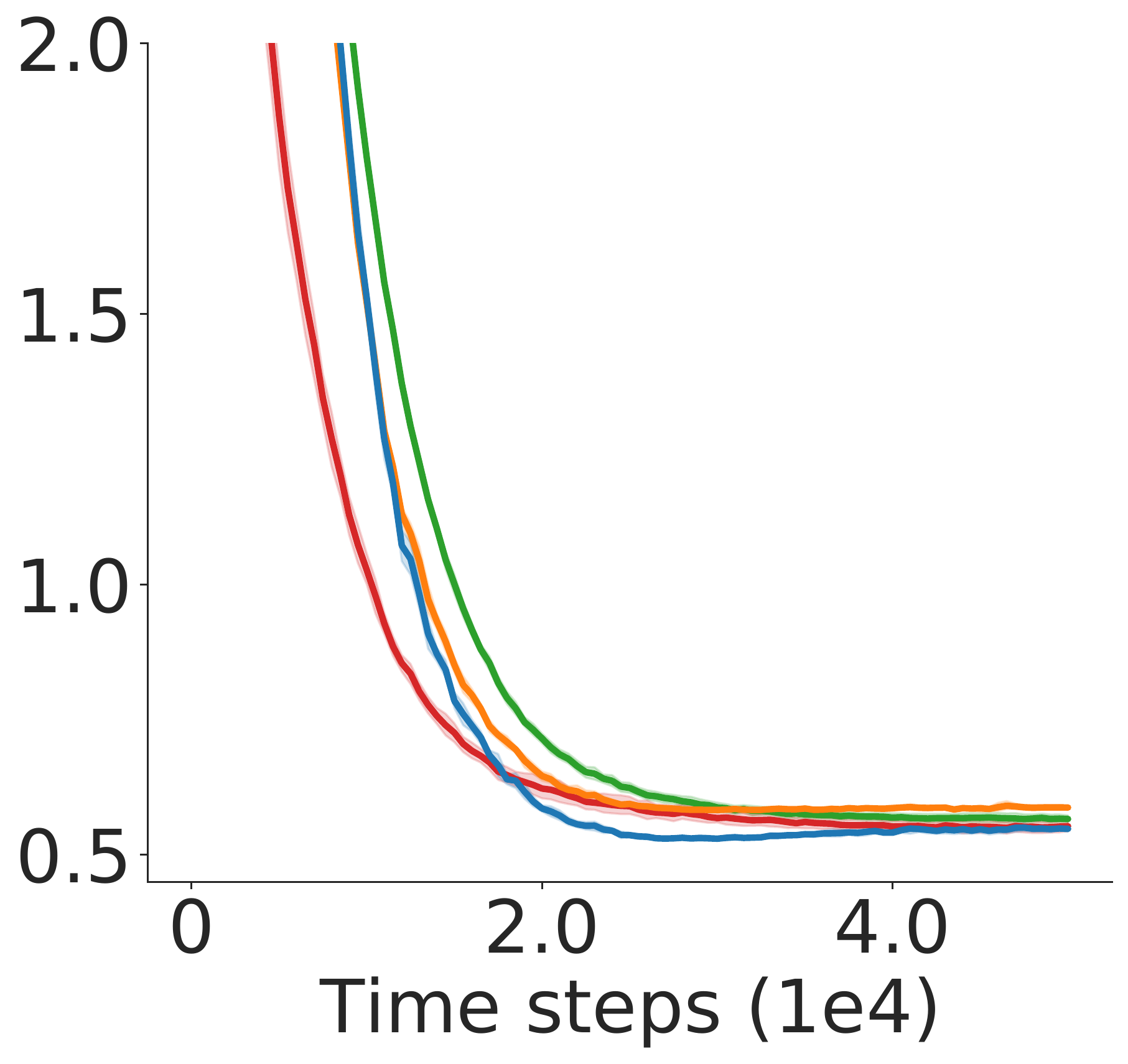} \\
    
    {\tiny Final Buffer Training} & {\tiny Final Buffer Testing} 
    &{\tiny Expert Training} & {\tiny Expert Testing} 
    &{\tiny Medium Training} & {\tiny Medium Testing} \\
    \end{tabular}
    \caption{Offline policy evaluation in different continuous control environment. Repeat for 10 different seeds for network initialization. Each row represent a different environment: Pendulum-Angle, HalfCheetah-Vel and Ant-Dir. Each column represent training or testing MSE in different offline dataset. All the x-axis is the training time steps and y-axis is the average MSE.}
    \label{fig:ope_result}
\end{figure}

We demonstrate the effectiveness of our newly designed operator networks in both offline policy evaluation and policy optimization.
$\rdist$ is an uniform distribution on a set of rewards known before hand.
We test both the performance of the value function on the training rewards as well as a set of new testing rewards.

We mainly compare our different designs of operator networks with successor feature \citep{barreto2017successor} in offline policy evaluation case.
To make fair comparison, each time the reward sampler will give all the training rewards value: for successor representation, the multiple training rewards serve as basis functions; for operator q-learning, these serve as multiple times of randomly sampled picked from the reward sampler.
Since our setting is for offline, it is not fair to compare with Generalized Policy Improvement(GPI) in the policy optimization case.

\subsection{Reward Transferring for Offline Policy Evaluation}
\myparagraph{Environment and Dataset Construction}
We conduct experiments on three continuous control environments: Pendulum-Angle, HalfCheetah-Vel and Ant-Dir;
Pendulum-Angle is an environment adapted from Pendulum, a classic control environment whose goal is to swing up a bar and let it stay upright. 
We modify its goal into swinging up to a given angle, sampled randomly from a reward sampler.
Ant-Dir and HalfCheetah-Vel are standard meta reinforcement learning baseline adapted from \citet{finn2017model}.

We use online TD3 \citep{fujimoto2018addressing} to train a target policy $\pi$ on the original predefined reward function for a fixed number of iterations.
The offline dataset is collected by either: 1) the full replay buffer of the training process of TD3, or 2) a perturbing behavior policy with the same size , where the behavior policy selects actions randomly with probability $p$ and with high exploratory noise $\N(0, \sigma^2)$ added to the remaining actions followed \citet{fujimoto2019off}. See Appendix~\ref{sec:exp_appendix} for more details of the construction of offline dataset and the designs of training and testing reward functions for different environments. 

\myparagraph{Criteria}
We evaluate the zero-shot reward transfer ability of operator network during the training process, where we feed both the training reward functions (the fixed random sampled training rewards we use for training) and the unseen random testing reward functions into our operator network, and evaluate the performance of our predicted value function $\G_\theta[r]$ with the true $q_{\pi,r}$ with mean square error(MSE) metric:
$$
\mathrm{MSE} = \frac{1}{n_{\mathrm{test}}}\sum_{i=1}^{n_{\mathrm{test}}} (q_{\pi,r}(y_i) - \G_\theta[r](y_i))^2\,,
$$
where $\{y_i\}_{i=1}^{n}$ are the state-action pairs whose states are drawn from the initial distribution of the environment, and $q_{\pi,r}(y_i)$ is the ground truth action value function computed from the trajectory drawn from target policy $\pi$. 
Notice that TD3 provides a deterministic policy and the dynamics is also deterministic, so we just need to collect one trajectory for each initial $y_i$.
For all the plots, we report the average MSE for multiple training and testing rewards.

\myparagraph{Results}
Figure~\ref{fig:ope_result} shows the comparison results with {\tt Successor Feature}, {\tt Attention-based Operator} network, {\tt Linear-based Operator} network and {\tt Vanilla Operator} network.
We can see that the Attention-based operator network achieves a much better initialization thanks to its self-normalized nature, thus it converges faster than any other methods in all setups.  
It is worth to mention that in the extreme case when the offline dataset is close to the target policy, the initialization (almost equal weight) of Attention-based structure can achieve even better performance than the one after training (HalfCheetah Final Buffer and Expert Behavior).
Compared with successor feature, which needs linear assumption to guarantee generalization, all the operator networks achieve more stable generalization performance in testing reward transferring.

\newcommand{\newfigsize}{0.3}
\begin{figure}
    \centering
    \includegraphics[width =.95\linewidth]{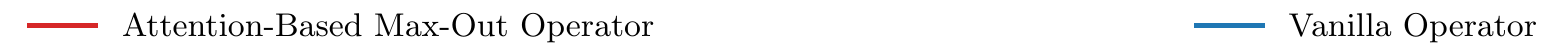}
    \begin{tabular}{ccc}
    \raisebox{0.5em}{\rotatebox{90}{\small Episodic Reward}}
    \includegraphics[width = \newfigsize\textwidth]{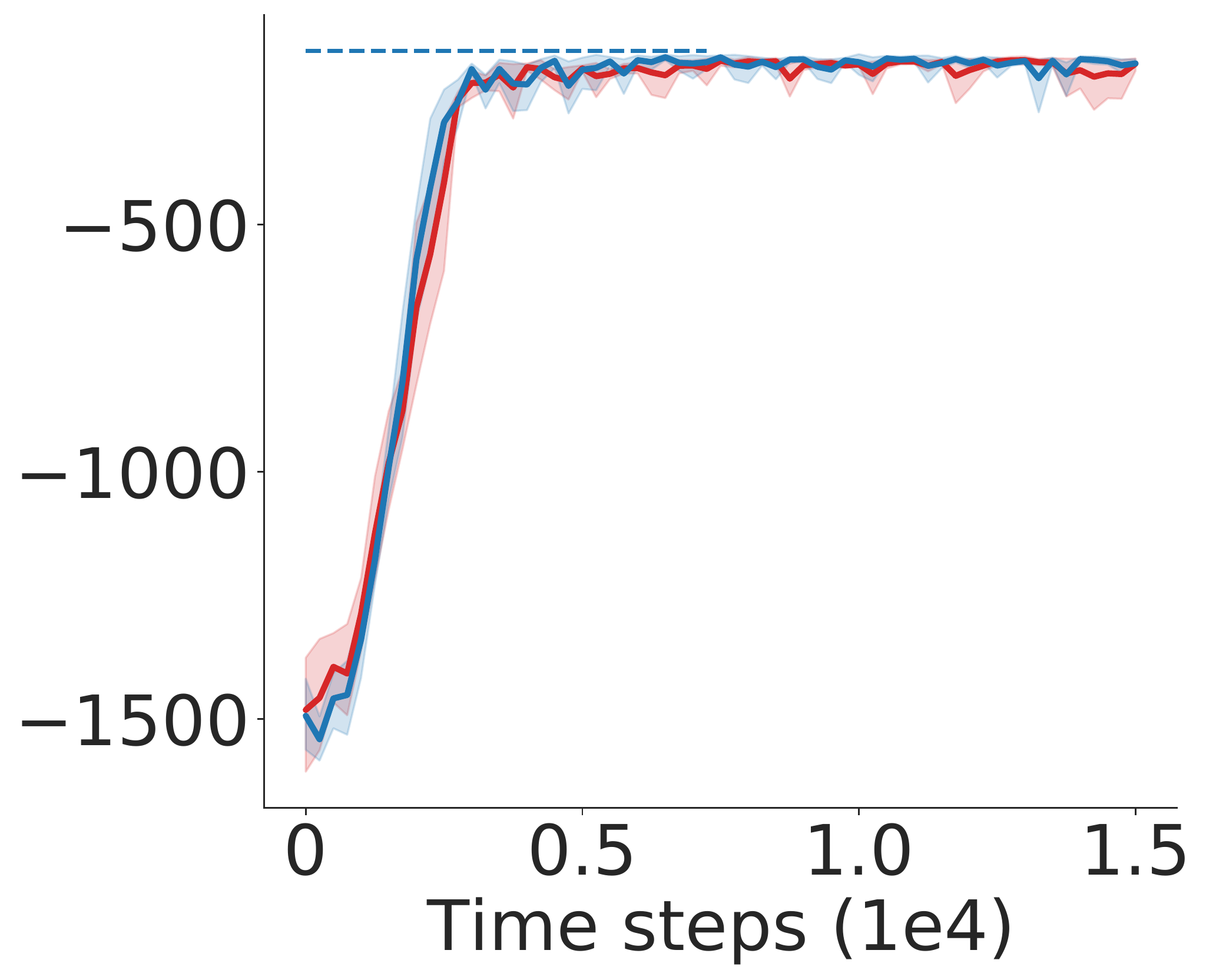} &
    \includegraphics[width = \newfigsize\textwidth]{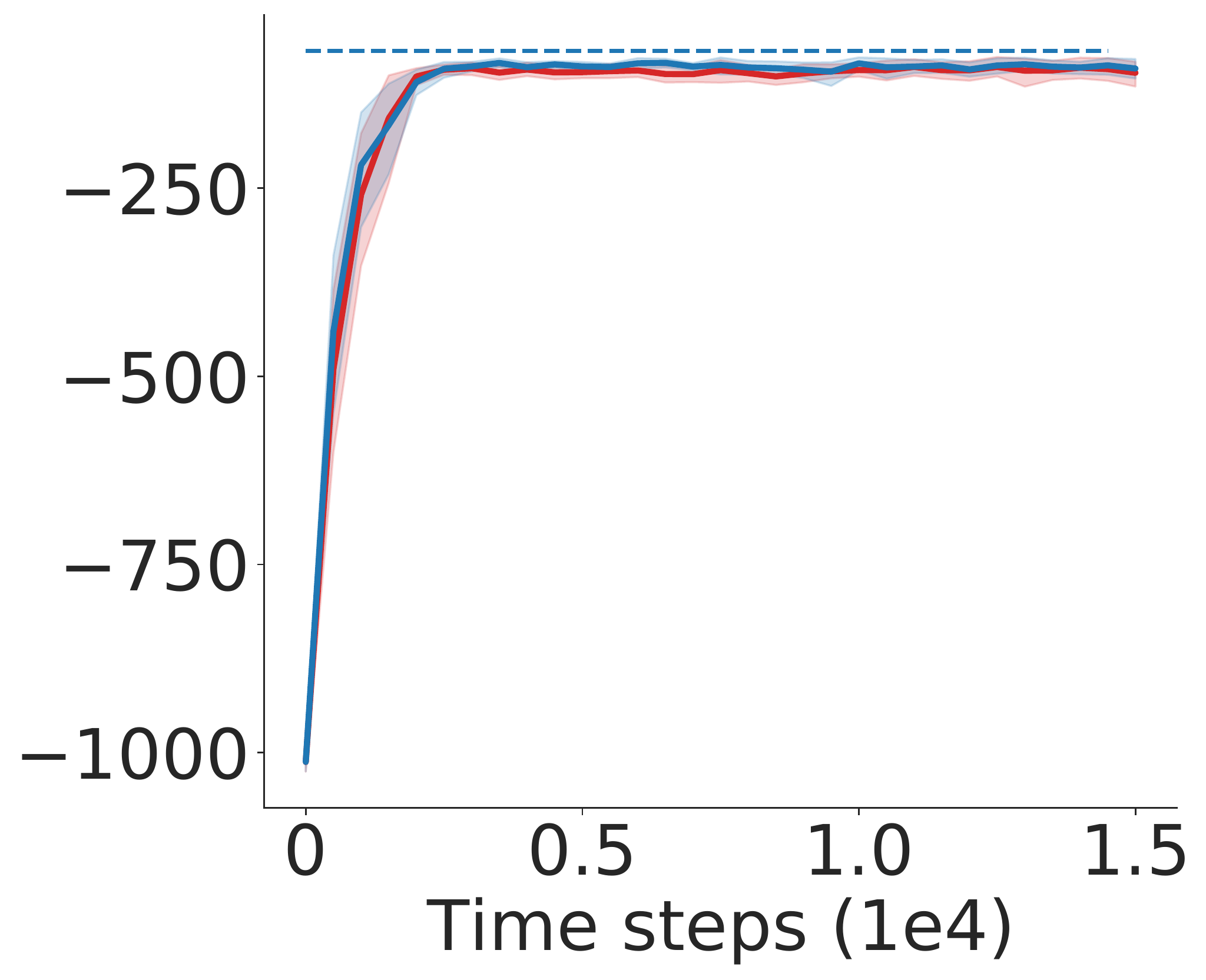} &
    \includegraphics[width = \newfigsize\textwidth]{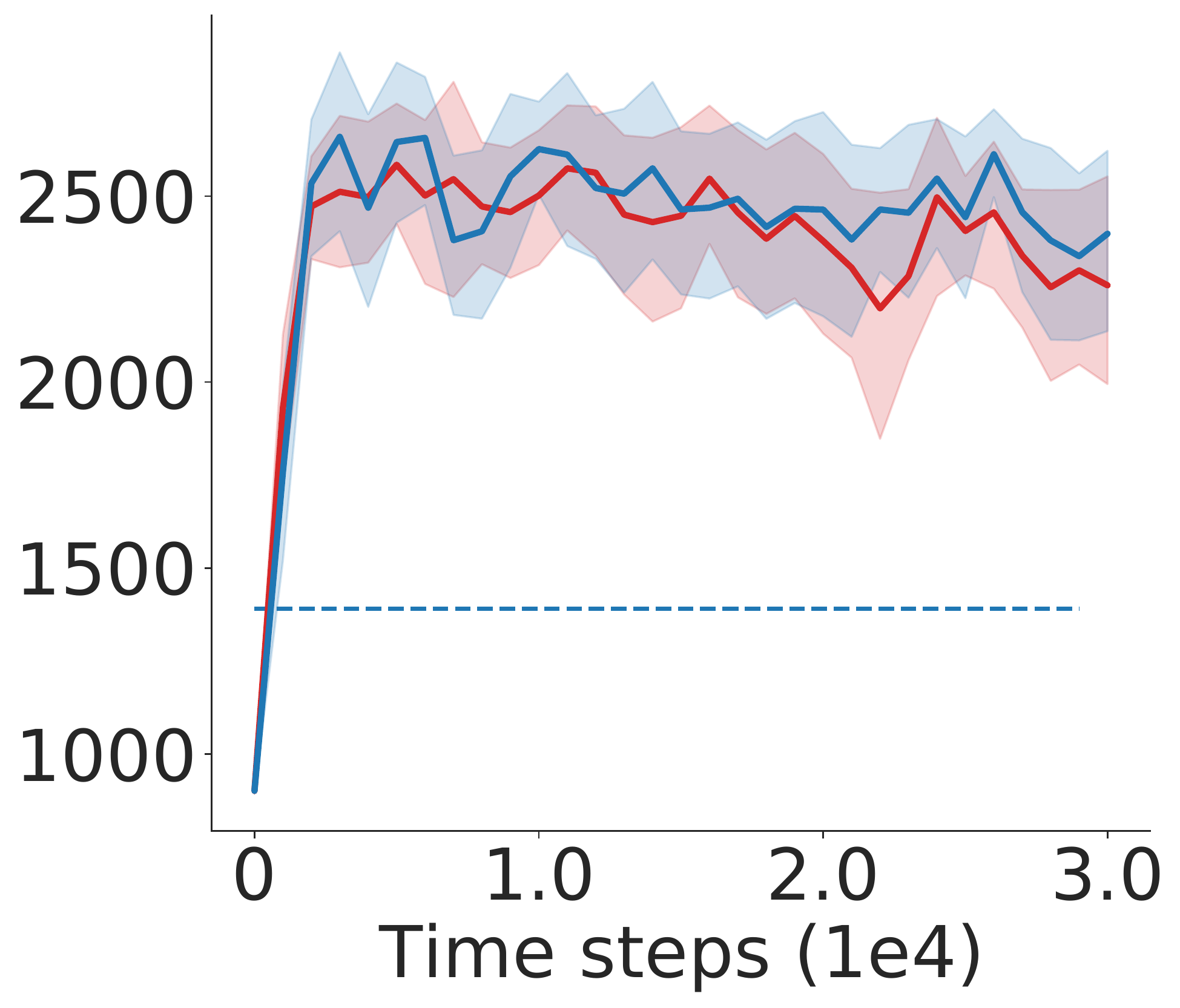}\\
    {\small (a) Pendulum-Angle}  & {\small (b) Half-CheetaVel} & {\small (c) Ant-Dir}\\
    \end{tabular}
    \caption{Offline policy optimization in different continuous control environment. Repeat for 10 different seeds for network initialization. All the x-axis is the training time steps and y-axis is the episodic rewards for the predefined testing reward function.}
    \label{fig:opi_result}
\end{figure}

\subsection{Reward Transferring for Offline Policy Optimization}
We compare the (Attention-based) {\tt Max-out Operator} and the {\tt Vanilla Operator}
on policy optimization tasks. 
All the task settings and the offline dataset collection processes 
are exactly the same as previous section.
However, since the dataset is constructed mainly by one trained target policy for a single predefined reward function, it is impossible to perform well on transferring to another reward with offline training due to distribution shift, thus we only pick the predefined reward function as our testing reward.
For Pendulum-Angle environment, we discretized the action domain and perform standard Operator DQN; for other two high dimension domains, it is impossible to discretize the action space, so we modified Batch Constrained deep Q-learning\citep{fujimoto2019off} into an operator version; see Appendix~\ref{sec:exp_appendix} for more details.

\myparagraph{Criteria}
We evaluate the zero-shot reward transfer ability of the operator network, where we evaluate the episodic reward on the current policy (for discrete action we pick the maximum of the critic value as policy) if we feed the operator network with the test reward.

\myparagraph{Results}
Both operator networks can achieve almost the same performance of the converged best target policy trained online with TD3 even in a zero-shot transferring.
For Ant-Dir, the TD3 has not converged to optimum policy, the offline optimization can even outperform the target policy using the same replay buffer.

%% file: tex/appendix.tex
\clearpage
\onecolumn
\appendix

\section{Proof}
\label{sec:appendix_proof}

\subsection{Proof of Proposition~\ref{pro:property_gpi}}
\begin{proof}
For any reward function $r$, we have:
\begin{align*}
    \Gpi[r](x) =& \E_{\tau \sim \pi} 
    \left [\sum_{t=0}^{\infty} \gamma^t r(s_t, a_t)|x_0 = x \right ] \\
    =& \sum_{t=0}^{\infty} \gamma^t \Ppi^t[r](x) \\
    =& \left(\sum_{t=0}^{\infty} \gamma^t \Ppi^t\right)[r](x) \\
    =& \left(I-\gamma\Ppi\right)^{-1} [r](x)\,.
\end{align*}

For the properties, we just need to prove that $\Ppi$ is linear, monotonic and invariant to constant, then $\Gpi = \sum_{t=0}^{\infty} \gamma^t \Ppi^t$ would automatically satisfies all the properties.

By the definition of $\Ppi$, we have:
\begin{enumerate}
    \item Linearity:
    \begin{align*}
        \Ppi[\alpha r_1 + \beta r_2](x) =& \E_{x'\sim p_\pi(\cdot|x)}[\left(\alpha r_1 + \beta r_2\right)(x')] \\
        =& \alpha\E_{x'\sim p_\pi(\cdot|x)}[r_1(x')] + \beta \E_{x'\sim p_\pi(\cdot|x)}[r_2(x')] \\
        =& \alpha\Ppi[r_1](x) + \beta\Ppi[r_2](x).
    \end{align*}
    \item Monotonicity: From linearity, we only need to prove $\Ppi[\Delta](x) \geq 0,~\forall x\in \X$, where
    \begin{align*}
        \Ppi[\Delta](x) = \E_{x'\sim p_\pi(\cdot|x)}[\Delta(x')] \geq 0
    \end{align*}
    \item Invariant to constant function. From the definition of $\Ppi$ we know:
    \begin{align*}
        \Ppi[r_C](x) = \E_{x'\sim p_\pi(\cdot|x)}[r_C(x')] = C
    \end{align*}
\end{enumerate}
\end{proof}

\subsubsection{Proof of Theorem~\ref{thm:approximate_gpi}}
\begin{proof}
Since $r$ is a continous function, for any $\varepsilon > 0$, there exists $\delta > 0$ such that 
$|r(x) - r(y)|\leq \varepsilon$ for $d(x,y)\leq \delta$.

From Eq.~\eqref{eqn:point_evaluation} we have: 
$$\Gpi[r](x) = \frac{1}{1-\gamma}\int_{\X}d_\pi(x'|x)r(x')dx'.$$

Since the domain $\X$ is compact, there exists a sufficiently large $m$ so that we can cover the domain with $m$ balls $\X = \cup_{j=1}^m B_j$ with $B_j$ as a $\delta-$radius ball centered at $\xi_j$:
$$
B_j = \{x\in \X~|~d(\xi_j,x)\leq \delta\}.
$$

Define $L_j$ as the non-overlapping Voronoi tessellation induced by $\xi_j$:
$$
L_j = \{x\in \X~|~ d(x,\xi_j) \leq d(x,\xi_i),~\forall i\neq j\},
$$
we have:
\begin{align*}
    \Gpi[r](x) =& \frac{1}{1-\gamma}\int_{\X}d_\pi(x'|x)r(x')dx' \\
    =& \frac{1}{1-\gamma}\sum_{j=1}^m \int_{L_j} d_\pi(x'|x) r(x') dx' \\
    \in& \frac{1}{1-\gamma}\sum_{j=1}^m w_\pi(L_j|x) (r(\xi_j) \pm \varepsilon) \\
    =& \frac{1}{1-\gamma}\left(\sum_{j=1}^m w_\pi(L_j|x) r(\xi_j)\right) \pm \varepsilon\,,
\end{align*}
where $w_\pi(L_j|x) = \int_{L_j} \frac{d_\pi(x'|x)}{1-\gamma} dx'$ as the conditional probability mass on $L_j$. The inequality leverages the property of the continuous function that any $x'\in L_j$ has $d(x',\xi_j)\leq \delta$ which implies $|r(x') - r(\xi_j)| \leq \varepsilon$.

Notice that we can approximate $w_\pi(L_j|x)$ with any universal function approximator, and $|r(\xi_j)|$ is bounded by $1$, we can safely says that we can approximate $\Gpi[r](x)$ with $\G_\theta[r](x)$ with $\varepsilon$ accuracy.
\end{proof}

\subsection{Proof of Proposition~\ref{pro:property_gmax}}
\begin{proof}
Eq.~\eqref{eqn:maxpi} is immediate from the definition of $q_{*,r}$.
To prove the properties, we can leverage the properties of $\Gpi$:
\begin{enumerate}
    \item Linearity: For any $r_1, r_2$ we have:
    \begin{align*}
        \Gmax[r_1 + r_2](x) =& \max_\pi \Gpi[r_1 + r_2](x) \\
        =& \max_\pi \{\Gpi[r_1](x) + \Gpi[r_2](x)\} \\
        \leq & \max_\pi \{\Gpi[r_1](x)\} + \max_\pi \{\Gpi[r_2]\} \\
        =& \Gmax[r_1] + \Gmax[r_2].
    \end{align*}
    For any $\alpha\in \RR$ we have
    $$\Gmax[\alpha r](x) = \max_\pi \Gpi[\alpha r](x) = \alpha \max_\pi \Gpi[r](x) = \alpha \Gmax[r](x).$$
    \item Monotonicity: Suppose the optimum policy for $r$ is $\pi_{*,r}$, we have:
    \begin{align*}
        \Gmax[r](x) = \G_{\pi_{*,r}}[r](x) \leq \G_{\pi_{*,r}}[r+\Delta](x) \leq \max_\pi \Gpi[r+\Delta](x) = \Gmax[r+\Delta].
    \end{align*}
    \item Invariant to constant function:
    \begin{align*}
        \Gmax[\alpha r + r_C](x) =& \max_\pi \{\Gpi[\alpha r + r_C](x)\} \\
        =& \max_\pi \{\alpha\Gpi[r](x) + C/(1-\gamma)\} \\
        =& \alpha \max_\pi \{\Gpi[r](x)\} + C/(1-\gamma) \\
        =& \alpha \Gmax[r](x) + C/(1-\gamma).
    \end{align*}
\end{enumerate}

\end{proof}

\clearpage
\section{The Adjoint View of Linear Operator}
\label{sec:adjoint}

We take a deeper look into the operator $\Gpi$.
In Proposition \ref{pro:property_gpi}, we know $\Gpi$ is a linear operator.
It is not hard to show under $\L_{\infty}$ space with $\ell_\infty$ norm, $\Gpi$ is a bounded operator:
$$
    \sup_{x\in \X} \Gpi [r](x) \leq \sup_{x\in \X} r(x) / (1 - \gamma)\,.
$$
Notice that any bounded linear operator has its corresponding adjoint operator.
It turns out the adjoint view of $\Gpi$ is closely related to the recent density-based methods in OPE \citep{tang2019doubly, nachum2020reinforcement, uehara2020minimax, jiang2020minimax}. 

\myparagraph{Koopman Operator and Transfer Operator}
To introduce the adjoint operator, we first briefly review the transfer (Perron–Frobenius) operator and its adjoint Koopman operator, which is important in control theory, stochastic process and RL. 
\begin{mydef}\label{def:operator}
For a transfer probability density function $p(y|x)$ in domain $\X$, we define its Koopman operator $\P$ and transfer operator $\P^\dag$ for any $\mu \in \L_1(\X)$, $f\in \L_{\infty}(\X)$ as
\begin{align*}
    \P[f](x) &= \int_{\X} p(y|x) f(y) dy\,, \\ 
    \P^\dag[\mu](y) &= \int_{\X} p(y|x) \mu(x) dx\,.
\end{align*}
\end{mydef}
The transfer operator $\P^\dag$ is also called "forward operator" while the Koopman operator $\P$ is called "backward operator", as they are the solution operators of the forward (Fokker–Planck) and backward Kolmogorov equations\citep{lasota2013chaos}, respectively.
It is easy to show that $\P$ and $\P^\dag$ are adjoint to each other
\begin{align*}
\langle \P^\dag[\mu], f\rangle = \langle \mu, \P[f]\rangle.
\end{align*}
In particular, if we pick $\mu = \delta_x$ as a delta measure, we have:
$$
    \langle p(\cdot|x), f\rangle = \P[f](x) = \langle \delta_x, \P[f]\rangle = \langle \P^\dag[\delta_x], f\rangle,
$$
which implies $p(\cdot|x) = \P^\dag[\delta_x]$.

\myparagraph{Koopman Opeartor in RL}
In RL, our interest of transfer kernel is $p_\pi(x'|x) = p(s'|s,a)\pi(s'|a')$.
The corresponding operators $\Ppi$ and $\Ppistar$ satisfy the following lemma:

\begin{lem}
(\citet{tang2019doubly} Lemma A.2) The two operators $\Ppistar$ and $\Ppi$ are adjoint:  
$
    \langle \Ppistar[\mu], f\rangle = \langle \mu, \Ppi[f] \rangle\,,
$
and the corresponding Bellman equations can be written in an operator view
\begin{align}
    & q_{\pi,r} = r + \gamma \Ppi [q_{\pi,r}]\,, \notag\\
    & d_{\pi}(\cdot|x) = (1-\gamma)\delta_x + \gamma \Ppistar [d_{\pi}(\cdot|x)]\,, \label{eqn:bellman_dpi}
\end{align}
where $d_{\pi}(\cdot|x)$ is defined in Eq.~\eqref{eqn:def_dpi}. 
\end{lem}

The above Lemma gives an adjoint view of the Bellman equation.
In particular, if we consider the adjoint operator of $\Gpi$, from Proposion~\ref{pro:property_gpi} we have:
$$
    \Gpi^\dag = \left(\sum_{t=0}^\infty \gamma^t \Ppi^t\right)^\dag 
    =\sum_{t=0}^\infty \gamma^t  \left(\Ppi^\dag\right)^t = (I-\gamma \Ppistar)^{-1}\,,
$$
and similar to the Bellman equation for $d_\pi$ in Eq.~\eqref{eqn:bellman_dpi}, we have:
$$
   d_\pi(\cdot|x)/(1-\gamma) = \Gpistar[\delta_x] = \delta_x + \gamma \Ppistar[\Gpistar[\delta_x]]\,.
$$
In the adjoint view, our design of $w_\theta$ can be seen as mapping any delta measure to a set of weighted delta measure as:
$$
\Gpistar[\delta_x] \approx \frac{1}{1-\gamma}\sum_{j=1}^m   w_\theta(\xi_j~|~x) \delta(x' = {\xi_j})\,,
$$
which can be viewed as an importance sampling based estimator similar to many recent OPE methods \citep{liu2018breaking}.

\clearpage
\section{Discussion of Different Designs}
\label{sec:discuss_design}

\subsection{Time Complexity of Different Designs of Evaluation Operator}
We mentioned two different design of $w_\theta(\xi_j|x)$ in Section~\ref{sec:design}, one is attention based design in Eq.~\eqref{eqn:attention_design_gpi}
$$
    w_\theta({\xi_j}|x) = \frac{\exp(f_{\theta_f}(\xi_j)^\top g_{\theta_g}(x))}{\sum_{k=1}^m \exp(f_{\theta_f}(\xi_k)^\top g_{\theta_g}(x))}\,,
$$
and another one is the linear decomposition design in Eq.~\eqref{eqn:linear_design_gpi}
$$
    w_\theta({\xi_j}|x) = f_{\theta_f}(\xi_j)^\top g_{\theta_g}(x)\,.
$$

One caveat of attention based design is the time complexity.
Suppose in each iteration we need to do gradient descend on a batch of $b$ desired points $x_i$, to compute $b$ different $\G_\theta[r](x_i)$, we need to evaluate $O(bm)$ number of $w(\xi_j|x_i)$ for $b$ $x_i,~\forall i\in [b]$ and $m$ different $\xi_j,~\forall j\in [m]$.

However, if we are using linear decomposition design, we can achieve a faster time complexity with $O(b+m)$, where we can decompose $\G_\theta$ as:
\begin{align}\label{eqn:linear_decompose_gpi}
    \G_\theta[r](x) =& \frac{1}{1-\gamma}\sum_{i=1}^m w(\xi_j|x) r(\xi_j)\,,\notag \\
    =& \left( \frac{\sum_{i=1} r(\xi_j) f_{\theta_f}(\xi_j)}{1-\gamma}\right)^\top g_{\theta_g}(x)\,,
\end{align}
where $\frac{\sum_{i=1} r(\xi_j) f_{\theta_f}(\xi_j)}{1-\gamma}$ can be firstly computed in time $O(m)$ with $m$ reference poitns $\{\xi_j\}_{j\in [m]}$, and we can reuse it to compute for $\G_\theta[r](x_i),~\forall i\in [b]$, with a total time complexity as O(m+b).

\subsection{Practical Attention-Based Design}
To avoid the multiplicative time complexity, we need to find a way to reduce the time complexity for attention-based design.

\myparagraph{Reduce the number of reference points}
One approach is to reduce the number of reference points $m$.
In practice, we find this simple solution is extremely useful and we can achieve almost the same running time as linear decomposition-based design.
In all the environments we test, we pick $128$ reference points uniformly random from the whole offline dataset $\D = \{x_i, r_i\}$ and fix it for later operator network structure.
It turns out that increasing the number of reference points can only have very small performance gain in general.
Although our operator network design is different from \cite{lu2019deeponet}, but the effect for the number of reference points is similar; see \citet{lu2019deeponet} for more quantitative discussion.

\myparagraph{Approximate by random feature attention}
\citet{peng2021random} proposed a way to approximate attention network by random feature. 
The approximation structure will eventually become a linear decomposition model which can reduce the time complexity to $O(b+m)$.
In this way, we can incorporate more reference points into the network design.
We will leave this to future work once the codebase of random feature attention in \citet{peng2021random} is available.

\subsection{Connection with General Policy Improvement(GPI)}
Our max-out architecture is similar to General Policy Improvement(GPI) \citep{barreto2018transfer} where we all consider a max-out structure in reward transferring.
However,
GPI does not learn its policies and its corresponding successor features directly from the offline dataset, but the policies and its corresponding successor features are picked from previous tasks with different rewards.
For example, if we are given $m$ pretrain tasks with reward $r_1, r_2,\ldots, r_m$, GPI will train $m$\footnote{It can be less than $m$ if the max-out of previous policies/successor features already yields a good performance} differnt policies $\pi_i$ where $\pi_i$ is the trained policy specifically for reward $r_i$ which is trained in a online manner.
Compared with GPI, our different $K$ copies of $w_{\theta_k}$ is randomly initialed and jointly trained in a offline manner. Since it is just served as part of the network structure, there is no exact meaning for each of the $w_{\theta_k}$.

\clearpage
\section{More Related Works}
\myparagraph{Universal Value Function and Successor Features}
The notion of \textit{reward transfer} is not new in reinforcement learning, and has been studied in literature.
Existing methods aim to capture a \textit{concept} that can represent a reward function.
By leveraging the concept into the design of the value function network, the universal value function can generalize across different reward functions.
Different methods leverage different concepts to represent the reward function.
Universal value function approximators (UVFA) \citep{schaul2015universal} considers a special type of reward function that has one specific goal state, i.e. $r_g(s,a) = f(s,a, g)$,
and leverage the information of goal state into the design;
Successor features (SF) \citep{barreto2017successor,barreto2018transfer,borsa2018universal, barreto2020fast} considers reward functions that are linear combinations of a set of basis functions, i.e. $r = w^\top \phi$,
and leverage the coefficient weights $w$ into the design.
Both methods rely on the assumption of the reward function class to guarantee generalization.
And typically they cannot get access to the actual concept directly, 
and need another auxiliary loss function to estimate the concept from the true reward value \citep{kulkarni2016deep}.
Our method is a natural generalization on both methods and can directly plug in the true reward value directly.

\myparagraph{Multi-objective RL, Meta Reinforcement Learning and Reward Composing}
Multi-objective RL \citep[e.g.][]{roijers2013survey,van2014multi, li2019multi, yu2020gradient} 
deals with learning control policies to simultaneously optimize over several criteria.
Their main goal is not transferring knowledge to a new unseen task, but rather cope with the conflict in the current tasks.
However, if they consider a new reward function that is a linear combination of the predefined criteria functions \citep{yang2019generalized}, e.g. lies in the optimal Pareto frontiers of value function, then it can be viewed as a special case of SF, which is related to our methods.

Meta reinforcement learning \citep[e.g.,][]{duan2016rl, finn2017model, nichol2018first, xu2018meta, rakelly2019efficient, Zintgraf2020VariBAD} can be seen as a generalized settings of reward transfer, where the difference between the tasks can also differ in the underlying dynamics. 
And they usually still need few-shot interactions with the environment to generalize, differ from our pure offline settings.

Works on reward composing \citep{van2019composing, tasse2020boolean} propose to compose reward functions in a boolean way. However, their setting is a special MDP where the reward functions of interest only differ at the absorbing state sets.

\myparagraph{Reward Free RL}
Recent works on reward (task) free RL \citep[e.g.][]{wang2020reward,jin2020reward, zhang2020task} break reinforcement learning into two steps: exploration phase and planning phase.
In the exploration phase, they don't know the true reward functions and only focus on collecting data by exploration strategy.
In the planning phase, they receive a true reward function and based on the data collected in the exploration phase, 
Our method can be seen as an intermediate phase in between to help reward transfer in the planning phase, where the zero-shot transfer can serve as a good initial during planning phase.

\myparagraph{Off Policy Evaluation(OPE)}
Our design of resolvent operator $\Gpi$ is highly related to the recent advances of density-based OPE 
methods \citep[e.g.,][]{liu2018breaking, nachum2019dualdice, tang2019doubly,mousavi2019black, zhang2020gendice,zhang2020gradientdice}, see more discussion in Section~\ref{sec:adjoint}.
However, density-based OPE methods usually focus on a fixed initial distribution 
while our conditional density in Eq.~\eqref{eqn:def_dpi} can be more flexible to handle arbitrary initial state-action pairs.
And for value-based methods, such as Fitted Q-Evaluation(FQE) \citep[e.g,][]{voloshin2019empirical}, though empirically better than density-based ones \citep{fu2021benchmarks}, usually cannot handle multiple reward functions simultaneously.

\clearpage
\section{Experimental Details}
\label{sec:exp_appendix}

\subsection{Reward Design}

\myparagraph{Pendulum-Angle}
The original reward function for Pendulum environment can be written as: 
$$
r(s,a) = -(\theta^2 + 0.1 * \mathrm{vel}^2 + 0.001 * \mathrm{force}^2)\,,
$$
where $\theta$ is the angle of the bar, $\mathrm{vel}$ is the angular velocity and the action is the angular force,
and the observation $s = [\cos(\theta), \sin(\theta), \rm{vel}]^\top$.

To change it into a multi reward environemnt, we consider:
\begin{align*}
    &r(s,a|\theta_0)= -((\theta-\theta_0)^2 + 0.1 * \rm{vel}^2 + 0.001 * \rm{force}^2)\,,
\end{align*}
where in training phase, we randomly sample $32$ training rewards uniformly from:
$$
    \F_{train} = \{r(\cdot|\theta_0): \theta_0\in [-0.4 * \pi, 0.4 * \pi]\},
$$
and in testing phase, we randomly sample $16$ testing rewards uniformly from:
$$
    \F_{test} = \{r(\cdot|\theta_0): \theta_0\in [-0.6 * \pi, 0.6 * \pi]\}.
$$
Notice that $\F_{train}$ does not cover $\F_{test}$, our design aims to see generalizability of different methods.

\myparagraph{HalfCheetah-Vel}
HalfCheetah-Vel is adapted from \citet{finn2017model}, where the goal is to achieve a target velocity running forward.
The reward function followed exactly as \citet{rakelly2019efficient} codebase as:
$$
   r(s,a|v_{target}) = -1.0 * |v-v_{target}| - 0.05 * \|a\|_2^2.
$$
where $v$ is the average velocity in the x-axis and $a$ is the vector of action.
For training tasks, our $32$ target velocities $v_{target}$ are sample uniformly random from $[0.7, 1.3]$; and for testing tasks, our $16$ target velocities are sample uniformly from $[0.6, 1.4]$.

\myparagraph{Ant-Dir}
Ant-Dir is adapted from \citet{rakelly2019efficient} codebase\footnote{\url{https://github.com/katerakelly/oyster/tree/master/rlkit/envs}}, where the goal is to keep a 2D-Ant moving in a given direction.
The reward function can be written as:
$$
    r(s,a|\theta) = (v_x * \cos(\theta) +  v_y * \sin(\theta))
    - 0.5 * \|a\|_2^2 - 0.005 * \rm{contact~cost} + 1.0,
$$
where the contact cost can be computed using the state, and the $1.0$ is the survival reward to prevent the ant to suicide at the initial training.
For training tasks, we sample $32$ $\theta$ randomly from $[-\frac{\pi}{4},\frac{\pi}{4}]$ and for testing we sample $16$ $\theta$ randomly from $[-\frac{\pi}{3},\frac{\pi}{3}]$.

\subsection{Offline Dataset Construction Details}
For offline dataset construction, we exactly follow \citet{fujimoto2019off} and collect the offline dataset as (1) final replay buffer during training the target policy using TD3 or (2) sample from a behavior policy follow the target policy but with probability $p$ to select actions randomly and with an exploratory noise $\N(0,\sigma^2)$ added to the action in the other $1-p$ probability. See \citet{fujimoto2019off} database\footnote{\url{https://github.com/sfujim/BCQ/tree/master/continuous_BCQ}} for more details.

For the hyper-parameter of $p$ and $\sigma$, see Table~\ref{tab:offline_construction} for more details for each environment.

\begin{table}[h]
    \centering
    \begin{tabular}{|c|c|c|c|}
    \hline 
        & Dataset size $n$ & Random Action Probability $p$ & Noise variance $\sigma$ \\\hline 
        Pendulum-Angle Experts & 2e4 & 0.1 & 0.3 \\\hline 
        Pendulum-Angle Medium & 2e4 & 0.3 & 0.3\\\hline 
        HalfCheetah-Vel Expert& 1e5& 0.1 & 0.1\\\hline 
        HalfCheetah-Vel Medium& 1e5& 0.3 & 0.1\\\hline 
        Ant-Dir Expert &2e5 & 0.1 & 0.1\\\hline 
        Ant-Dir Medium &2e5 & 0.3 & 0.1\\\hline 
    \end{tabular}
    \caption{Parameter of behavior policy for different environments.}
    \label{tab:offline_construction}
\end{table}

\subsection{Modified Operator BCQ in Mujoco Environment}
In high dimension environment such as HalfCheetah-Vel and Ant-Dir, it is impossible to discretize the action space and apply operator DQN.
Thus we implement an actor-critic style policy optimization adapted from the current state of the art offline policy optimization method BCQ\citep{fujimoto2019off}, where the actor is a combination of an imitated actor $g_w:\Sset\to\Aset$ and a perturbation network $\zeta_\phi$.
We use exactly the same encoder-decoder structure as our imitation network $g_w$ as BCQ;
for the perturbation network, since now it depends on the different reward function, we use a high dimention vanilla operator network to form the mapping from
$r$ to $\zeta$ as an operator network, where the final action $a$ is sampled from
$$
    a \sim \rm{Clip}(\zeta_\phi[r](s,g_w(s)) + g_w(s)).
$$
This can be a naive extension using vanilla operator network for reward transfer in BCQ.
See our code in supplementary material for more details.

Since the implementation of combining BCQ is not the main purpose of our paper, we haven't tried other methods yet, so we think there is still a large room to improve in future.
\subsection{Other Details}

\myparagraph{Training Hyper-parameter}
The hyper-parameters are summarized in Table~\ref{tab:hyper_parameter},
where we pick Adam as our optimizer, and for all tasks we set the learning rate $\varepsilon$ as $0.001$, and the target network update rate $\alpha$ is $0.005$ and the batch size is $256$.
All attention-based method we fixed the number of reference points $m = 128$.
And we set the Max-out repetition $K$ in policy optimization as $8$.
\begin{table}[h]
    \centering
    \begin{tabular}{cc}
    \hline
        Hyper-parameter & Value \\
    \hline
        Optimizer & Adam \\
        Learning Rate $\varepsilon$ & 0.001\\
        Target Network Update Rate $\alpha$ & 0.005 \\
        Batch Size & 256  \\
        Number of reference points $m$ for Attention & 128 \\
        Max Out Size $K$ & 8\\
    \hline
    \end{tabular}
    \caption{Hyper-parameter for implementation.}
    \label{tab:hyper_parameter}
\end{table}

\myparagraph{Training Speed}
Each offline experiment takes approximate $5-30$ minutes with one RTX 2080 ti GPU depending on environment and task with $15000 - 50000$ total training steps.